\documentclass{article}

\usepackage{iclr2027_conference,times}

\usepackage{amsmath,amsfonts,bm}

\def\eqref#1{equation~\ref{#1}}

\def\1{\bm{1}}

\DeclareMathAlphabet{\mathsfit}{\encodingdefault}{\sfdefault}{m}{sl}
\SetMathAlphabet{\mathsfit}{bold}{\encodingdefault}{\sfdefault}{bx}{n}

\usepackage{algorithm}
\usepackage{algpseudocode}
\usepackage{hyperref}
\usepackage{url}
\usepackage{booktabs}
\usepackage{multirow}
\usepackage[table]{xcolor}
\usepackage{tabularx}
\usepackage{graphicx}
\usepackage{wrapfig}
\usepackage{needspace}
\usepackage{capt-of}
\usepackage[above,below]{placeins}
\usepackage{tcolorbox}
\usepackage{listings}
\usepackage{enumitem}
\usepackage{pifont}
\tcbuselibrary{breakable}
\definecolor{ragagentgreen}{RGB}{235,247,237}

\definecolor{accentgreen}{RGB}{28,112,72}     
\definecolor{tableheadgreen}{RGB}{242,248,244} 
\usepackage[capitalize,nameinlink,noabbrev]{cleveref}
\crefname{equation}{Eq.}{Eqs.}
\Crefname{equation}{Eq.}{Eqs.}
\crefname{appendix}{Appendix}{Appendices}
\crefname{algorithm}{Algorithm}{Algorithms}
\crefname{table}{Table}{Tables}
\crefname{figure}{Figure}{Figures}
\crefname{section}{Section}{Sections}
\crefformat{appendix}{Appendix~#2#1#3}
\newcommand{\pcref}[1]{\cref{#1}}
\newcommand{\pointmark}[1]{\textcolor{accentgreen}{\ding{#1}}}
\newcommand{\obs}[1]{\textcolor{accentgreen}{\textbf{#1}}}
\newcommand{\key}[1]{\textbf{#1}}
\newlist{pointlist}{itemize}{1}
\setlist[pointlist]{leftmargin=1.6em, itemsep=0.5pt, topsep=1.5pt, parsep=0pt,
  partopsep=0pt, label={}}
\setlist[itemize]{leftmargin=1.4em, itemsep=1pt, topsep=2pt, parsep=0pt,
  partopsep=0pt}

\definecolor{coreboxbg}{RGB}{243,250,245}
\newtcolorbox{corebox}{
  colback=coreboxbg,
  colframe=accentgreen!80!black,
  boxrule=0.45pt, arc=1.5pt,
  left=6pt, right=6pt, top=2.5pt, bottom=2.5pt,
  before skip=4pt, after skip=4pt,
  breakable
}

\definecolor{questionbg}{RGB}{243,250,245}
\definecolor{questionframe}{RGB}{128,170,148}
\newtcolorbox{questionbox}{
  colback=questionbg,
  colframe=questionframe,
  boxrule=0.4pt, arc=1pt,
  left=6pt, right=6pt, top=2.5pt, bottom=2.5pt,
  before skip=4pt, after skip=4pt,
  breakable
}

\definecolor{prompttitlegreen}{RGB}{225,241,231}
\definecolor{promptbodygreen}{RGB}{249,252,250}
\definecolor{promptbordergreen}{RGB}{128,170,148}
\newtcolorbox[auto counter,number within=section]{promptbox}[2][]{
  breakable,
  width=\linewidth,
  colback=promptbodygreen,
  colbacktitle=prompttitlegreen,
  colframe=promptbordergreen,
  coltitle=black,
  fonttitle=\bfseries,
  fontupper=\normalsize,
  boxrule=0.5pt,
  titlerule=0.4pt,
  arc=1.2mm,
  boxsep=0pt,
  left=8pt,right=8pt,top=6pt,bottom=6pt,
  toptitle=5pt,bottomtitle=5pt,
  before skip=8pt,after skip=10pt,
  title={#2},
  title after break={#2 (continued)},
  #1
}

\definecolor{casegray}{RGB}{247,248,249}
\definecolor{casegraytitle}{RGB}{235,238,241}
\definecolor{casegrayborder}{RGB}{175,182,189}
\newtcolorbox{casepanel}[2][]{
  breakable,
  width=\linewidth,
  colback=casegray,
  colbacktitle=casegraytitle,
  colframe=casegrayborder,
  coltitle=black,
  fonttitle=\small\bfseries,
  fontupper=\small,
  fontlower=\small,
  boxrule=0.5pt,
  titlerule=0.4pt,
  arc=1.2mm,
  boxsep=0pt,
  left=7pt,right=7pt,top=5pt,bottom=5pt,
  toptitle=4pt,bottomtitle=4pt,
  before skip=6pt,after skip=6pt,
  before upper={\setlength{\parindent}{0pt}\setlength{\parskip}{3pt}},
  title={#2},
  title after break={#2 (continued)},
  #1
}
\tcbset{casesuccess/.style={
  colback=promptbodygreen,
  colbacktitle=prompttitlegreen,
  colframe=promptbordergreen
}}
\lstdefinestyle{casepython}{
  language=Python,
  basicstyle=\ttfamily\small,
  keywordstyle=\bfseries,
  commentstyle=\itshape\color{black!65},
  columns=fullflexible,
  keepspaces=true,
  showstringspaces=false,
  breaklines=true,
  aboveskip=2pt,belowskip=0pt,
  tabsize=4
}

\title{CoMem: Learning to Anticipate Failures\\from Retrieved Experience}

\author{%
Yun Shen\textsuperscript{1} \quad
Ming Shen\textsuperscript{2,3} \quad
Tianjun Yao\textsuperscript{2,*} \\
\normalfont \textsuperscript{1}The Ohio State University \quad
\textsuperscript{2}Shenzhen Loop Area Institute \quad
\textsuperscript{3}Shanghai Jiaotong University \\
\normalfont \textsuperscript{*}Corresponding author.
}

\iclrfinalcopy

\begin{document}

\ClearShipoutPictureBG

\maketitle
\lhead{Preprint}

\begin{abstract}
Reflection lets LLM agents improve from their own attempts, but the memory
it builds is local to a single task, so an agent reflecting on its own tends
to revisit the same explanation for failure. Cross-task reflective memory
addresses this in two forms: retrieved memory keeps past experience explicit
and looks up related cases, which is specific but bounded by the bank, and
parametric memory internalizes experience into parameters, which generalizes
but is grounded in nothing beyond the task description. When both are used,
existing agents combine them additively, feeding their outputs to the solver
side by side, so retrieved experience never changes what the learned memory
produces. We introduce \textbf{CoMem} (Compounding Memory), which makes
retrieved experience an input to the learned memory itself, at training and
at inference. The agent builds a bank of its own successful and failed
reflections, learns from retrieved cases to write failure hypotheses,
candidate explanations of how an attempt is likely to go wrong, and at test
time reasons from newly retrieved cases to guide each attempt. Varying the
retrieved cases gives the memory different perspectives, yielding
reflective diversity that is not merely sampled but also grounded in retrieved experience. Across two
backbones and five benchmarks in math, code, and multi-hop QA, CoMem
achieves strong overall performance. Additional evaluations show that it
generalizes to out-of-distribution tasks and enables smaller models to
provide effective guidance to stronger solvers.
\end{abstract}

\section{Introduction}
\label{sec:introduction}

Large language models (LLMs) now solve a wide range of reasoning and
coding tasks~\citep{openai2026gpt6astra,qwen2026qwen38max,deepseek2026v41flash},
and a natural next step is agents that keep improving from their own
experience rather than from a stronger external
model~\citep{zelikman2022star,madaan2023selfrefine}. At test time, the
dominant mechanism for such improvement is reflection, in which the
agent verbalizes what went wrong after a failed attempt and carries this
guidance into the next trial~\citep{shinn2023reflexion}. Reflection-based
frameworks have become a standard component of language agents for
programming, mathematical reasoning, and multi-hop question
answering~\citep{yao2023react,madaan2023selfrefine,shinn2023reflexion}.

\begin{figure}[t]
  \centering
  \includegraphics[width=\linewidth]{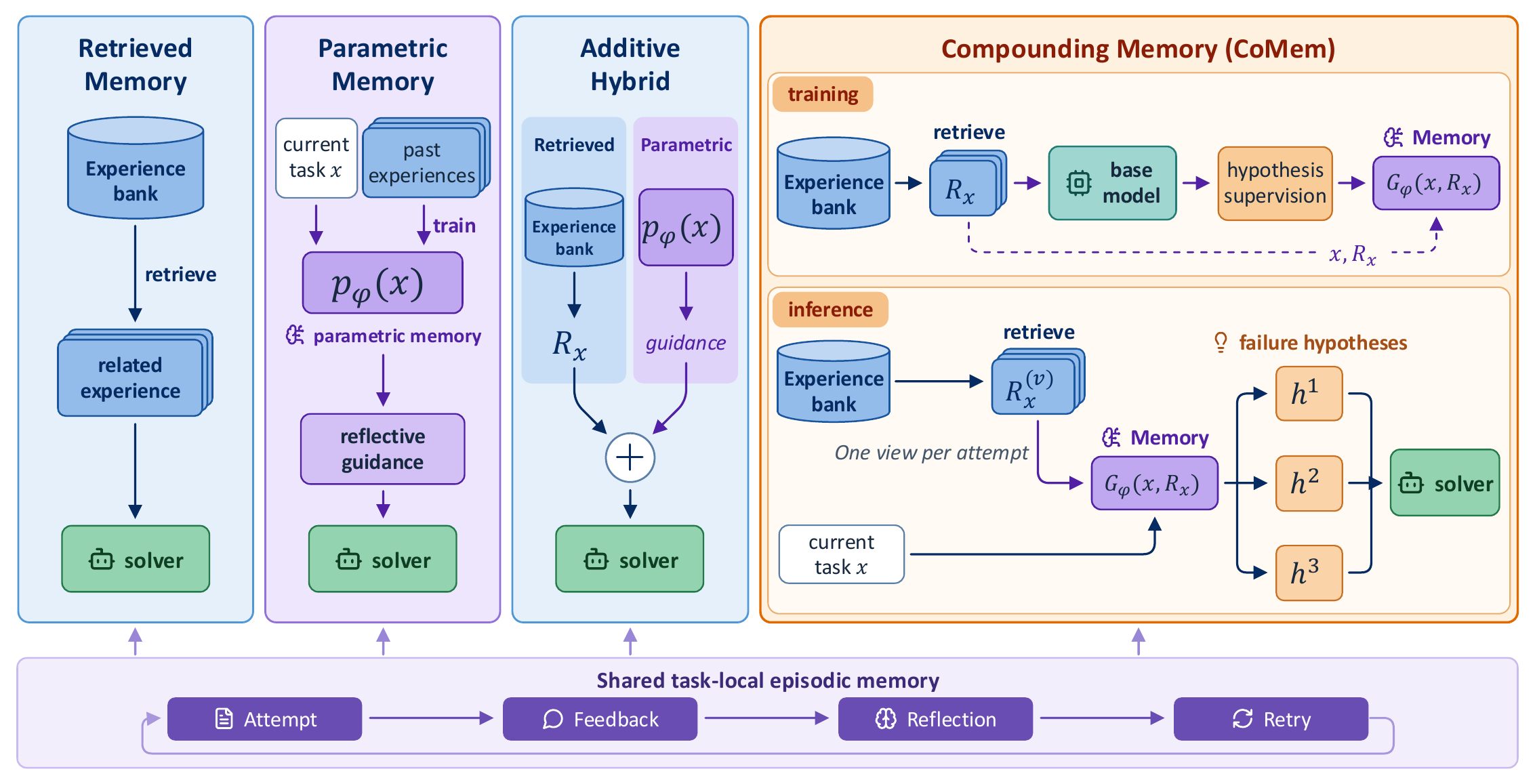}
  \caption{\textbf{Reflective memory designs and CoMem.}
  Retrieved memory recalls past experience, parametric memory generates
  guidance from the task alone, and additive hybrids combine both.
  CoMem uses retrieved experience to construct supervision and train an
  experience-conditioned generator. At inference, different retrieved pairs
  elicit diverse failure hypotheses.}
  \label{fig:comem-overview}
\end{figure}

The effectiveness of reflection, however, is bounded by its diversity.
Studies of self-correction report that a model left to reflect on its own
tends to produce repetitive or superficial diagnoses that miss the actual
cause of failure~\citep{huang2024selfcorrect,yao2024retroformer}, so
later attempts revisit the same explanation instead of exploring
alternative ones. Conversely, more diverse reflections correlate strongly
with higher task success~\citep{lingam2025diversity,yao2026parammem}.
Existing work introduces diversity by varying prompts
and sampling~\citep{lingam2025diversity}, searching over multiple
reasoning branches~\citep{yao2023treeofthoughts,zhou2023language},
retrieving trajectories of previously solved tasks from a cross-sample
memory bank~\citep{lingam2025diversity}, and, most recently, training a
lightweight parametric memory that generates reflective signals
from patterns internalized across many
tasks~\citep{yao2026parammem}. The latter two are notable because they
let an agent draw on experience beyond the current task, and both can be
built entirely from the base model's own trajectories.

Retrieval and parametric memory expand diversity through opposite
mechanisms, and each has a ceiling. Retrieval provides concrete,
task-specific experience, but its coverage is limited by the bank and
by what embedding similarity can surface. Learned embeddings capture
compositional structure poorly and tend to collapse into low-rank
subspaces, which narrows the set of
retrievable cases~\citep{nguyen2023generative,weller2025theoretical,guo2023embedding}.
Parametric memory generalizes beyond the bank but generates from the
task description alone. Its diversity comes only from sampling
temperature, with no concrete experience to ground each sample, so its
outputs can drift toward generic advice. Current systems place
the two side by side, concatenating retrieved trajectories and
parametric outputs in the solver's prompt~\citep{yao2026parammem}, which
treats remembered experience and learned experience as two independent
sources to be summed. We argue this misses where the diversity lies.
\begin{questionbox}
Can an agent's remembered experience and its learned experience be
made to compound, rather than merely add, so that reflection becomes
more diverse than either form of memory alone can make it?
\end{questionbox}

Our answer rests on a simple observation about how experience becomes
useful. Remembering a specific past case and internalizing a general
lesson are complementary parts of the same process. A
retrieved case is only as useful as the agent's ability to see what it
shares with the task at hand, and a learned lesson is only as specific as
the concrete cases it is anchored to. Diversity arises when the same
internalized knowledge is applied to different remembered cases,
yielding different and equally grounded conclusions. We introduce
\textbf{CoMem} (\textbf{Co}mpounding \textbf{Mem}ory), a reflective
memory in which the parametric and the retrieved are no longer separate
sources but one loop (\pcref{fig:comem-overview}). The agent builds a bank of its own successful and
failed reflections, learns from that bank how to read related past
experience, and at inference retrieves experience for each task
and reasons from it to produce failure hypotheses about how an attempt
might fail. These guide the first attempt, with different sets used
alongside accumulated reflections in subsequent attempts. \emph{Retrieval
supplies task-specific experience, while parametric memory provides the
ability to generalize from it.} Each retrieved case gives the memory a
different perspective, adding a new source of reflective diversity
grounded in the agent's own experience. Every step of the loop runs on the base model alone,
making CoMem fully self-improving.

We evaluate CoMem with two backbone models on five benchmarks spanning
mathematical reasoning, code generation, and multi-hop question
answering. It achieves strong overall performance, produces measurably more diverse failure
hypotheses than either frozen generation or retrieval-free self-training,
generalizes to out-of-distribution tasks and demonstrates weak-to-strong
transfer, with hypotheses from a smaller model improving a larger solver. Our
contributions are threefold:

\pointmark{172}~\key{Compounding memory.}
  We identify that existing agents treat retrieved and parametric
  reflective memory as additive sources, and propose CoMem, a
  compounding memory that trains the parametric component to reason
  from retrieved successful and failed experience, with all supervision
  generated by the base model itself.
\quad\pointmark{173}~\key{Compounding as a source of diversity.}
  We show that compounding is a distinct source of reflective
  diversity. CoMem generates more diverse failure hypotheses than either
  memory alone, and case studies illustrate how this broadens the errors
  an agent considers during problem solving.
\quad\pointmark{174}~\key{Strong empirical performance.}
  We demonstrate strong performance across two backbones and five
  benchmarks spanning three domains, together with effective out-of-distribution
  generalization and weak-to-strong transfer.

\section{Related Work}
\label{sec:related-work}

\textbf{Reflection-based agents and the diversity bottleneck.}
Self-Refine~\citep{madaan2023selfrefine} and Reflexion~\citep{shinn2023reflexion}
established the reflect-and-retry loop, in which an agent built on
ReAct~\citep{yao2023react} critiques a failed attempt in natural language and
conditions the next attempt on that critique; CRITIC~\citep{gou2023critic} and
Self-Debug~\citep{chen2024teaching} ground the critique in tool or execution
feedback, the setting we also adopt. The loop, however, saturates quickly.
Without external feedback, self-correction rarely helps and can
degrade performance~\citep{huang2024selfcorrect,kamoi-etal-2024-llms}; even with feedback, an
agent that has committed to an explanation seldom produces a new one, a
failure named Degeneration-of-Thought~\citep{liang-etal-2024-encouraging} and
observed as behavior collapse when self-correction is trained with
RL~\citep{kumar2024training}, with alignment-induced mode collapse as a likely
root cause~\citep{kirk2024understanding}. Remedies inject diversity from within
the current task: DoT~\citep{lingam2025diversity} reduces redundant
reflections, Self-Contrast~\citep{zhang-etal-2024-self-contrast} contrasts
multiple solving perspectives, SPP~\citep{wang-etal-2024-unleashing}
simulates several personas, Verbalized
Sampling~\citep{zhang2026verbalizedsamplingmitigatemode} elicits diverse
samples by prompt, ToT~\citep{yao2023treeofthoughts} and
LATS~\citep{zhou2023language} search over branches, and multi-agent
debate~\citep{du2023improvingfactualityreasoninglanguage,liang-etal-2024-encouraging}
pits agents against each other, though debate does not reliably beat simple
ensembling~\citep{smit2024goingmadlookmultiagent}. All of these resample or
search the current task. \emph{CoMem instead draws diversity from experience on
other tasks}, which brings us to memory.

\textbf{Memory for reflection: retrieved and parametric.}
Reflexion~\citep{shinn2023reflexion} keeps an episodic buffer of its own
reflections. Cross-sample
memory extends this to other tasks: DoT-bank~\citep{lingam2025diversity}
retrieves trajectories of similar solved problems,
ExpeL~\citep{zhao2024expel} distills past trajectories into reusable
insights, and Generative Agents~\citep{park2023generative} synthesize a
memory stream into higher-level reflections that are later retrieved. A
separate line stores experience in parameters:
Retroformer~\citep{yao2024retroformer} trains a reflector with policy
gradients. Retrieved memory is specific but bounded by the
bank and by embedding similarity; parametric memory generalizes but is
conditioned on the task alone. ParamAgent-plus~\citep{yao2026parammem}
augments parametric memory with cross-task trajectory retrieval.
Retrieved trajectories and generated reflections jointly guide the solver,
but the parametric memory does not use retrieved experience during training
or reflection generation. \emph{CoMem makes retrieved experience the input of
the parametric memory at both stages}, so that its supervision is
retrieval-grounded and each retrieved case yields a different failure
hypothesis.

\textbf{Self-improvement from self-generated experience.}
Training on a model's own outputs is a well-established route to
improvement: STaR~\citep{zelikman2022star} and
ReST\textsuperscript{EM}~\citep{singh2023beyond} filter self-generated
rationales by correctness and fine-tune the solver on them, and
SCoRe~\citep{kumar2024training} trains self-correction with multi-turn RL.
Closer to agents, Retroformer~\citep{yao2024retroformer} optimizes the
reflector against environment reward, and ExpeL~\citep{zhao2024expel} and
Agent Workflow Memory~\citep{wang2024agentworkflowmemory} distill insights
or workflows from an agent's own trajectories without training.
\emph{CoMem keeps the solver frozen and trains a memory on hypotheses generated
by the same backbone from its own retrieved successes and failures.}
The memory learns to relate these experiences to the current task,
producing guidance that can also transfer to stronger solvers.

\section{Preliminaries}
\label{sec:preliminaries}

\textbf{Reflection loop.}
Let $p_\theta$ be a language model that produces a solution $y$ for a task
$x$ and receives feedback $f$ (unit-test results, a correctness signal, or
an answer judge).

In Reflexion~\citep{shinn2023reflexion}, when attempt
$y_{k-1}$ fails, the model writes a reflection $r_{k-1}$ from $(x, y_{k-1},
f_{k-1})$, stores it in an episodic memory, and conditions the next attempt
on all reflections so far,
\begin{equation}
  y_k \sim p_\theta(\cdot \mid x, r_{1:k-1}).
  \label{eq:reflexion}
\end{equation}
Reflections are written after failure and by the same model that failed, so
across iterations they tend to restate one explanation.

\textbf{Cross-task memory.}
Two families of methods let the agent draw on experience from other tasks.

\textbf{Retrieved memory.}
Retrieval-based memory~\citep{lingam2025diversity} keeps a bank
$\mathcal{B}$ of trajectories from previously solved tasks and adds the
most similar ones to the prompt,
\begin{equation}
  y_k \sim p_\theta(\cdot \mid x, r_{1:k-1}, \textsc{Retrieve}(\mathcal{B}, x)).
  \label{eq:retrieval}
\end{equation}
\textbf{Parametric memory.}
Parametric memory~\citep{yao2026parammem} fine-tunes a lightweight module
$p_\phi$ on reflections written for training tasks and, at each iteration,
samples a generated reflection that is appended to the prompt,
\begin{equation}
  y_k \sim p_\theta(\cdot \mid x, r_{1:k-1}, r^g_k), \qquad
  r^g_k \sim p_\phi(\cdot \mid x).
  \label{eq:parametric}
\end{equation}
The generated reflection depends on $x$ alone, so different samples differ
only through temperature. When both memories are used, their outputs are
concatenated in the solver's prompt, and $p_\phi$ never sees what was
retrieved. \emph{CoMem, introduced next, replaces $p_\phi(\cdot \mid x)$ with a
memory that takes the retrieved experience as input.}

\section{CoMem: Compounding Retrieved and Parametric Memory}
\label{sec:method}

CoMem replaces the unconditioned generator $p_\phi(\cdot \mid x)$ of
\cref{eq:parametric} with a memory $G_\phi(\cdot \mid x, R_x)$ that reads
retrieved experience $R_x$ and writes failure hypotheses for the current
task. The memory is trained on supervision that is itself produced from
retrieved experience, so retrieval enters at both training and inference.
We describe the experience bank, the retrieval-grounded supervision used to
train $G_\phi$ (\pcref{sec:building}), hypothesis generation at inference
(\pcref{sec:generating}), and how the solver uses
the hypotheses (\pcref{sec:solving}). \cref{alg:comem-framework} in the appendix
summarizes the full procedure.

\subsection{Building the Memory}
\label{sec:building}

\textbf{Experience bank from the agent's own reflection.}
For every training task $x \in \mathcal{D}_{\mathrm{tr}}$ we run the base
model $p_\theta$ through the reflection loop of \cref{eq:reflexion} for
at most $K=3$ attempts, including the initial attempt. Tasks solved on the first attempt carry no
correction signal and are discarded. If a later attempt succeeds, the
transition from the failed attempt through its reflection to the repaired
solution, $\tau = (y_{k-1}, r_{k-1}, y_k)$, is stored as a positive entry in
$\mathcal{B}^{+}$. If the budget is exhausted, the final failed attempt and
its reflection are stored as a negative entry in $\mathcal{B}^{-}$. The
bank $\mathcal{B} = \mathcal{B}^{+} \cup \mathcal{B}^{-}$ therefore holds
only experience about how this model's attempts go wrong and, when a repair
was found, how it was found. Both types of entries are useful because a positive entry shows a
failure mode together with its fix, while a negative entry records a failure
that the model did not resolve within the attempt budget.
The complete contents of each memory entry are detailed in
\cref{app:implementation}.

\textbf{Retrieval-grounded supervision.}
To train $G_\phi$ we need, for each task, a target set of failure
hypotheses. Rather than obtaining these from a stronger model, we let
$p_\theta$ write them while reading its own related experience. For a
training task $x$, we embed the task statement and retrieve the top-$k$
most similar tasks from $\mathcal{B}^{+}$ and $\mathcal{B}^{-}$, excluding
$x$ itself. We then sample one entry from each shortlist according to
the softmax probabilities computed from its similarity scores.
The pair $R_x = (\tau^{+}, \tau^{-})$ is placed in the prompt
together with $x$ and an instruction to predict how an attempt at $x$ is
likely to fail, transferring relevant failure modes from the references
while accounting for differences between the tasks. The base model then
writes a hypothesis set $z_x \sim p_\theta(\cdot \mid x, R_x)$. We keep
candidates that contain concrete, task-specific items and discard generic
advice. The result is a supervision set
$\mathcal{S} = \{((x, R_x), z_x)\}$ in which every target is grounded in
two retrieved cases. The supervision prompts are provided in
\cref{app:pitfall-label-prompt}.

\textbf{Training.}
We fine-tune $p_\theta$ with LoRA~\citep{hu2022lora} on $\mathcal{S}$ to
obtain $G_\phi$ by minimizing the conditional negative log-likelihood
\begin{equation}
  \mathcal{L}(\phi) = -\sum_{((x, R_x), z_x) \in \mathcal{S}}
  \log G_\phi(z_x \mid x, R_x).
  \label{eq:sft}
\end{equation}
The retrieved pair is also included in the training input to $G_\phi$,
conditioning hypothesis generation on both the current task and retrieved
experience. To keep the memory from specializing to
the easiest tasks, which dominate the unbalanced supervision set, we balance
$\mathcal{S}$ through stratified sampling. Details of training-data sampling
are provided in \cref{app:datasets}.

\subsection{Generating Failure Hypotheses}
\label{sec:generating}

For a test task $x$, we retrieve from the same bank $\mathcal{B}$ and let
the memory write
\begin{equation}
  h \sim G_\phi(\cdot \mid x, R_x), \qquad R_x \sim
  \textsc{Retrieve}(\mathcal{B}, x).
  \label{eq:hypothesis}
\end{equation}
\begin{corebox}
\key{\cref{eq:hypothesis} is where the two memories compound.} The
distribution over hypotheses is a mixture over retrieved pairs,
$\sum_{R} \Pr(R \mid x)\, G_\phi(h \mid x, R)$.
Its diversity comes from sampling within a fixed pair, as in parametric
memory, and from \emph{variation across pairs}, which parametric memory alone
does not have.
\end{corebox}
We exploit the second source directly. For each test
task, we retrieve the top-$k$ entries from $\mathcal{B}^{+}$ and
$\mathcal{B}^{-}$ as in training, apply softmax to each shortlist's
similarity scores, and sample $V$ pairs with replacement by independently
drawing one entry from each shortlist according to these probabilities.
Generating one hypothesis set per pair yields $V$ views
$\{h^{(1)}, \dots, h^{(V)}\}$ of how the task might fail, each conditioned
on its corresponding retrieved pair.
$G_\phi$ learns to adapt retrieved experience to the current task,
allowing it to generate task-specific hypotheses even when the retrieved
cases are only loosely related.

\subsection{Solving with Failure Hypotheses}
\label{sec:solving}

We use the frozen base model $p_\theta$ as the solver and vary only the
guidance it receives. The first attempt is conditioned on the first hypothesis
set, and each subsequent iteration receives the next hypothesis set together with
the accumulated reflections,
\begin{equation}
  y_1 \sim p_\theta(\cdot \mid x, h^{(1)}), \qquad
  y_k \sim p_\theta(\cdot \mid x, r_{1:k-1}, h^{(k)}), \quad k > 1.
  \label{eq:solve}
\end{equation}
Hypotheses thus act before the first failure, so the agent can avoid an
error it has not yet made, and they rotate across iterations to encourage
reflections to consider alternative explanations on subsequent attempts.
During this reflection loop,
the solver receives only the memory's conclusions from the retrieved
training trajectories.

Following DoT-bank~\citep{lingam2025diversity}, we additionally maintain a
test-time bank $\mathcal{C}$ of trajectories solved during evaluation.
Tasks that remain unsolved after the loop in \cref{eq:solve} enter a
second solving stage, where the most similar successful trajectory from
$\mathcal{C}$ is added to the prompt alongside the failure hypotheses.
Newly solved tasks are added
to $\mathcal{C}$ immediately. This stage is shared with the DoT-bank
baseline, so any difference in the comparison comes from the hypotheses.

\textbf{Why compounding helps.}
The design connects these simple components by feeding retrieved experience
into the parametric memory at every stage. \key{Relative to retrieval alone,}
the agent receives a hypothesis that is specific to $x$ even when no
retrieved case matches it closely, and
does not need to fit those training trajectories into the solver's context.
\key{Relative to parametric memory alone,} every hypothesis is anchored to a real success or
failure, and the memory can be steered to a new region of failure modes by
changing what it reads rather than by raising the temperature. The whole
pipeline uses $p_\theta$ for the bank, for the supervision, for the memory,
and for the solver, so \emph{CoMem improves an agent using only that
agent's own experience}.

\section{Experiments}
\label{sec:experiments}

In this section, we evaluate CoMem on mathematical reasoning, code
generation, and multi-hop question answering. We compare its task performance
with reflection- and memory-based baselines and analyze the diversity of its
generated hypotheses. We further study the effect of training
data size, generalization to out-of-distribution tasks, and whether
hypotheses generated by a smaller model can improve a stronger solver.

\subsection{Experimental Setup}
\label{sec:experimental-setup}

\textbf{Datasets.}
We evaluate our method across three task domains.
For mathematical reasoning, we use MATH~\citep{hendrycks2021math}.
For code generation, we use HumanEval+~\citep{liu2023evalplus} and
LiveCodeBench~\citep{jain2025livecodebench}. For multi-hop question answering,
we use HotpotQA~\citep{yang2018hotpotqa} and
MuSiQue~\citep{trivedi2022musique}. For out-of-distribution (OOD) evaluation,
we use AIME24 and AIME25 as well as
OlympiadBench~\citep{he2024olympiadbench}
for mathematical reasoning,
BigCodeBench-Hard~\citep{zhuo2024bigcodebench} for code generation, and
2WikiMultiHopQA~\citep{ho2020twikimultihopqa} for multi-hop question
answering. Further dataset details are provided in \cref{app:datasets}.

\textbf{Evaluation metrics.}
We use accuracy for mathematical reasoning and multi-hop question
answering, and Pass@1 for code generation. For code tasks, generation and
reflection use only visible or synthetic tests. We evaluate the final program on hidden
tests and count it as correct only if it passes all tests. For MATH, we use
Math-Verify\footnote{\url{https://github.com/huggingface/Math-Verify}}
to check whether the predicted answer is mathematically equivalent to
the reference answer. For multi-hop QA, we use
Qwen3.6-35B-A3B~\citep{qwen36_35b_a3b} as a semantic judge to assess
answer correctness.
The judging prompt is provided in \cref{app:qa-judge-prompt}.
We report the mean and standard deviation over three seeds for all tasks.

\textbf{Models and baselines.}
We use Qwen3-4B and Qwen3-8B~\citep{yang2025qwen3} as backbone models,
both in non-thinking mode.
We compare our method against the following baselines:
\emph{Base}, which uses the corresponding backbone model to solve each
problem directly; Reflexion~\citep{shinn2023reflexion}, which uses
self-reflections stored in episodic memory to guide subsequent attempts;
DoT-Bank~\citep{lingam2025diversity}, which augments diverse self-reflections
with relevant successful trajectories retrieved from a cross-task memory
bank; and
ParamMem~\citep{yao2026parammem}, which learns cross-sample reflection
patterns through fine-tuning and generates reflections to guide the solver.
Throughout our experiments, the ParamMem baseline denotes ParamAgent-plus,
which augments parametric memory with cross-task trajectory retrieval.

\textbf{Implementation.}
Across all three domains, we fine-tune the memory for three
epochs with a learning rate of $2 \times 10^{-5}$ and a batch size of 128,
using LoRA with rank $r = 32$ and $\alpha = 64$.
For retrieval, we compute text embeddings using
\texttt{text-embedding-3-small}~\citep{openai_text_embedding_3_small}.
Further implementation details are
provided in \cref{app:implementation}.

\subsection{Experimental Results}
\label{sec:experimental-results}


\obs{Observation 1:} \textbf{CoMem outperforms strong baselines across diverse domains.}
\cref{tab:main-results} shows that CoMem achieves strong performance
across mathematical reasoning, code generation, and multi-hop QA.
CoMem improves over DoT-Bank and ParamMem, which already combines retrieved
and parametric memory. These gains suggest that conditioning hypothesis
generation on retrieved experience provides further benefits beyond their
additive combination. By learning to
interpret experiences from related tasks in the context of the current
problem, CoMem's memory can anticipate potential errors
beyond those revealed by the solver's own attempts. These additional
perspectives can complement self-reflection and help the solver reconsider
its approach across successive attempts.

On MATH, CoMem improves over ParamMem by 0.72 and 0.96 with Qwen3-4B
and Qwen3-8B, respectively. These gains build on already high baseline
accuracy, showing that CoMem can still provide useful additional guidance.
Similar tasks retrieved
from the memory bank may indicate errors the model could make
on the current task, helping hypothesis generation account for mistakes
that might otherwise be overlooked. Both coding benchmarks show larger
gains over ParamMem, reaching 3.28 and 1.43 on HumanEval+ and
1.14 and 3.23 on LiveCodeBench with Qwen3-4B and Qwen3-8B,
respectively. Here, hypotheses may complement test feedback by drawing
attention to potential implementation errors beyond those directly
indicated by the feedback. In multi-hop QA, CoMem matches ParamMem on
HotpotQA with Qwen3-4B and outperforms it on both benchmarks with Qwen3-8B.
Its lower MuSiQue accuracy with Qwen3-4B may stem from differences in
guidance. ParamMem~\citep{yao2026parammem} provides multiple solution
approaches, whereas CoMem's hypotheses generated by Qwen3-4B may not cover
a sufficiently broad range of reasoning errors.

\begin{table}[!t]
  \centering
  \footnotesize
  \setlength{\tabcolsep}{3pt}
  \renewcommand{\arraystretch}{1.30}
  \caption{Performance comparison of CoMem with baselines across three
  task domains using Qwen3-4B and Qwen3-8B as base models. We report accuracy
  for MATH and QA, and Pass@1 for code. Results are averaged over three
  random seeds, with standard deviations in superscript. All values are
  percentages. Bold marks the best result within each dataset and model.}
  \label{tab:main-results}
  \begin{tabularx}{\linewidth}{@{}ll*{5}{>{\centering\arraybackslash}X}}
    \toprule
    \rowcolor{tableheadgreen}
    Model & Method & MATH & HumanEval+ & LiveCodeBench & HotpotQA & MuSiQue \\
    \midrule
    \multirow{5}{*}{Qwen3-4B} & Base
      & $72.90^{\pm 0.74}$ & $73.42^{\pm 0.77}$ & $32.56^{\pm 0.31}$ & $74.6^{\pm 1.3}$ & $36.3^{\pm 0.3}$ \\
     & Reflexion
      & $85.73^{\pm 0.61}$ & $80.37^{\pm 1.00}$ & $43.67^{\pm 0.27}$ & $87.4^{\pm 0.5}$ & $63.5^{\pm 2.6}$ \\
     & DoT-Bank
      & $87.77^{\pm 0.29}$ & $85.28^{\pm 0.87}$ & $50.44^{\pm 0.96}$ & $91.3^{\pm 0.9}$ & $68.0^{\pm 2.8}$ \\
     & ParamMem
      & $92.33^{\pm 1.22}$ & $92.02^{\pm 1.62}$ & $53.53^{\pm 1.37}$ & $\mathbf{92.4^{\pm 0.2}}$ & $\mathbf{80.6^{\pm 1.0}}$ \\
     & \cellcolor{ragagentgreen}CoMem
      & \cellcolor{ragagentgreen}$\mathbf{93.05^{\pm 0.61}}$ & \cellcolor{ragagentgreen}$\mathbf{95.30^{\pm 0.29}}$ & \cellcolor{ragagentgreen}$\mathbf{54.67^{\pm 0.47}}$ & \cellcolor{ragagentgreen}$\mathbf{92.4^{\pm 0.6}}$ & \cellcolor{ragagentgreen}$78.6^{\pm 0.8}$ \\
    \midrule
    \multirow{5}{*}{Qwen3-8B} & Base
      & $79.38^{\pm 0.45}$ & $81.19^{\pm 0.58}$ & $41.00^{\pm 0.27}$ & $82.4^{\pm 0.2}$ & $43.1^{\pm 0.9}$ \\
     & Reflexion
      & $86.45^{\pm 0.17}$ & $84.87^{\pm 0.58}$ & $50.44^{\pm 1.03}$ & $92.6^{\pm 0.5}$ & $73.3^{\pm 0.9}$ \\
     & DoT-Bank
      & $90.17^{\pm 0.68}$ & $88.14^{\pm 0.29}$ & $53.89^{\pm 1.50}$ & $92.6^{\pm 0.5}$ & $73.5^{\pm 2.5}$ \\
     & ParamMem
      & $93.76^{\pm 0.17}$ & $93.66^{\pm 0.35}$ & $56.44^{\pm 1.07}$ & $93.6^{\pm 0.6}$ & $83.5^{\pm 2.0}$ \\
     & \cellcolor{ragagentgreen}CoMem
      & \cellcolor{ragagentgreen}$\mathbf{94.72^{\pm 0.94}}$ & \cellcolor{ragagentgreen}$\mathbf{95.09^{\pm 0.87}}$ & \cellcolor{ragagentgreen}$\mathbf{59.67^{\pm 0.27}}$ & \cellcolor{ragagentgreen}$\mathbf{94.9^{\pm 0.2}}$ & \cellcolor{ragagentgreen}$\mathbf{85.6^{\pm 3.1}}$ \\
    \bottomrule
  \end{tabularx}

\end{table}

\obs{Observation 2:} \textbf{CoMem enhances hypothesis diversity through retrieved reflections.}
We hypothesize that reflections from similar tasks
can suggest potential errors that the model might otherwise overlook,
leading to more diverse hypotheses for the current task. To study this,
we compare \emph{Frozen}, \emph{Vanilla SFT}, \emph{Frozen + Retrieval},
and \emph{CoMem}.
\emph{Frozen} generates hypotheses directly from the question without fine-tuning.
\emph{Vanilla SFT} fine-tunes a memory with LoRA on supervision targets
produced by the backbone. Neither configuration uses retrieved experience.
\emph{Frozen + Retrieval} adds examples retrieved from related tasks to the
frozen backbone's input without fine-tuning.
\emph{CoMem} includes examples retrieved from related tasks in the
memory's input prompt during both training and inference.
QA inputs include the provided context. All configurations generate 10
hypothesis sets per question at a sampling temperature of $T=1.0$.

\begin{wrapfigure}{R}{0.50\textwidth}
  \centering
  \includegraphics[width=\linewidth]{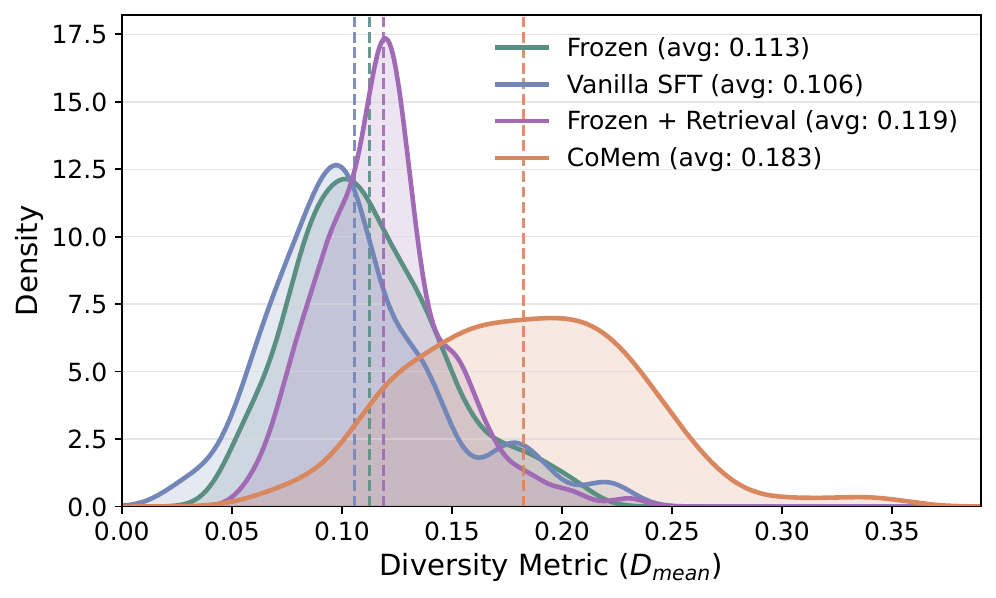}
  \caption{Hypothesis diversity on HumanEval+ with Qwen3-8B. Curves show
  $D_{\mathrm{mean}}$ distributions, with dashed lines marking means.}
  \label{fig:diversity-humaneval}
\end{wrapfigure}

For MATH and code generation, we evaluate Qwen3-4B and Qwen3-8B on MATH,
HumanEval+, and LiveCodeBench.
For each question, we embed the 10 generated
hypothesis sets and compute $D_{\mathrm{mean}}$ as the average cosine distance
between all pairs of embeddings. We then plot the distribution of
$D_{\mathrm{mean}}$ across questions, with higher values indicating
greater diversity. For QA, we evaluate Qwen3-8B on MuSiQue using
Self-BLEU-4~\citep{zhu2018texygen}, where lower scores indicate less overlap
among generated hypothesis sets, and Distinct-1/2~\citep{li2016diversity},
where higher scores indicate greater unigram and bigram diversity.


As shown in \cref{fig:diversity-humaneval}, Frozen + Retrieval and
Vanilla SFT remain close to Frozen on HumanEval+ with Qwen3-8B, while CoMem
substantially increases diversity. This advantage holds across mathematical
and programming tasks (\cref{app:diversity}). Across all four coding
settings, CoMem's gain over Frozen exceeds the sum of the two standalone
gains. This pattern supports compounding, as the memory learns to interpret
retrieved experience and derive more diverse failure hypotheses.

\begin{table}[!tb]
  \centering
  \small
  \setlength{\tabcolsep}{5pt}
  \renewcommand{\arraystretch}{1.30}
  \caption{Hypothesis diversity on MuSiQue with Qwen3-8B. Lower Self-BLEU-4
  and higher Distinct-1/2 indicate greater textual diversity. Bold marks
  the best result in each column.}
  \label{tab:qa-diversity}
  \begin{tabularx}{\linewidth}{l*{3}{>{\centering\arraybackslash}X}}
    \toprule
    \rowcolor{tableheadgreen}
    Method & Self-BLEU-4 $\downarrow$ & Distinct-1 $\uparrow$ & Distinct-2 $\uparrow$ \\
    \midrule
    Frozen & 0.598 & 0.149 & 0.414 \\
    Vanilla SFT & 0.508 & 0.172 & 0.412 \\
    \rowcolor{ragagentgreen}
    CoMem & \textbf{0.419} & \textbf{0.195} & \textbf{0.483} \\
    \bottomrule
  \end{tabularx}
\end{table}

The QA results provide further evidence of this effect
(\cref{tab:qa-diversity}). Although Vanilla SFT reduces overlap
relative to Frozen, its bigram diversity remains largely unchanged.
CoMem yields both lower Self-BLEU-4 and higher Distinct-1/2, indicating
less repetition and greater lexical variation in its hypothesis sets.
These findings suggest that incorporating experiences from related tasks
can diversify hypotheses by highlighting potential errors that may otherwise
be overlooked.
Case studies in \cref{app:case-study} illustrate how diverse
hypotheses can also diversify subsequent reflections, helping the model
recognize overlooked requirements and consider alternative solution strategies.

\begin{figure}[!ht]
  \centering
  \includegraphics[width=\linewidth]{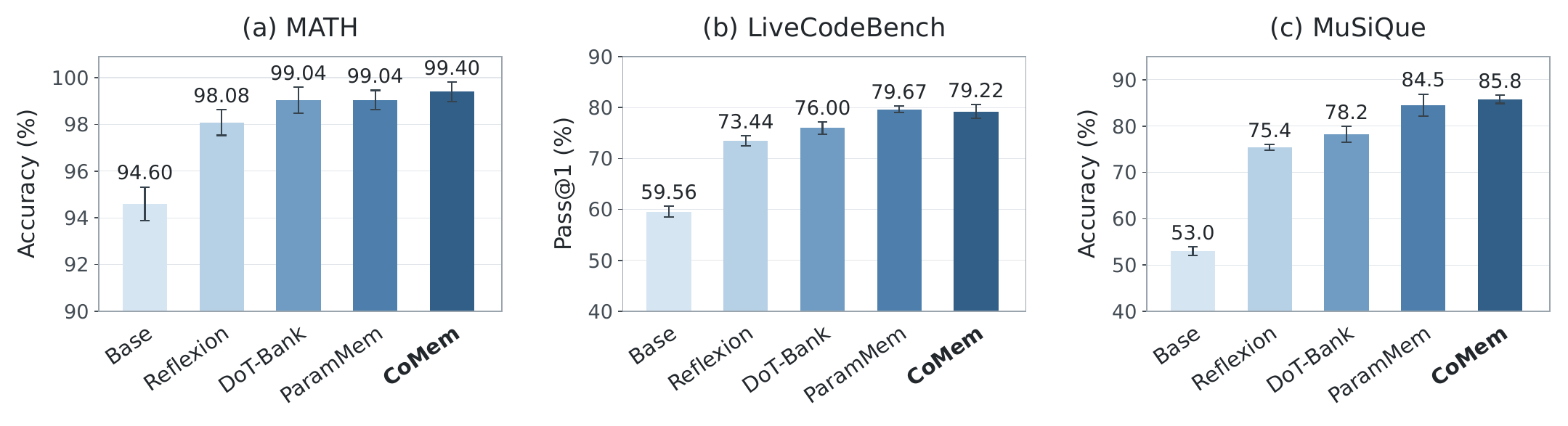}
  \caption{Weak-to-strong evaluation. We use Qwen3-Next-80B-A3B-Instruct
  as the task solver, guided by hypotheses generated by CoMem with
  Qwen3-8B. Bars show mean accuracy or Pass@1 over three random seeds.
  Error bars indicate one standard deviation.
  Compared with the baselines, the hypothesis-guided solver
  achieves the best or near-best performance across the three tasks.}
  \label{fig:weak-to-strong}
\end{figure}

\obs{Observation 3:} \textbf{Guidance from failure hypotheses transfers from smaller to stronger models.}
We investigate weak-to-strong transfer to determine whether hypotheses
generated by a smaller model can complement the reasoning of a stronger
solver. We use Qwen3-8B to generate hypotheses and provide them
to Qwen3-Next-80B-A3B-Instruct for problem solving on MATH,
LiveCodeBench, and MuSiQue. As shown in
\cref{fig:weak-to-strong}, CoMem achieves the best performance on
MATH, providing a further improvement even when accuracy is already
close to saturation. It also outperforms Reflexion and DoT-Bank on
LiveCodeBench and achieves the highest mean accuracy on MuSiQue,
showing that guidance from a smaller model remains useful to a stronger solver.

CoMem and ParamMem obtain similar results across these tasks. Some
hypotheses generated by the smaller model may concern errors that the
stronger solver already avoids, limiting the benefit of additional guidance.
The gains over DoT-Bank suggest that a smaller model can provide useful
guidance without matching the stronger solver's overall problem-solving
capability.


\obs{Observation 4:} \textbf{CoMem generalizes to out-of-distribution tasks.}
We evaluate whether a memory trained on source tasks can
provide useful guidance for out-of-distribution problems.
We use Qwen3-8B throughout. For mathematical reasoning, the memory
trained on MATH produces hypotheses for AIME24, AIME25, and OlympiadBench.
For programming, we use the APPS-trained
memory~\citep{hendrycks2021apps} to produce hypotheses that guide code
generation on BigCodeBench-Hard, evaluating transfer from algorithmic
programming problems to tasks that require combining function calls from
multiple libraries~\citep{zhuo2024bigcodebench}. For multi-hop QA, we
transfer the MuSiQue-trained memory to 2WikiMultiHopQA.
In each case, we use the source-trained memory to produce new hypotheses
for each OOD problem and provide these hypotheses to the solver.

\begin{table}[!t]
  \centering
  \footnotesize
  \setlength{\tabcolsep}{3pt}
  \renewcommand{\arraystretch}{1.30}
  \caption{Out-of-distribution evaluation with Qwen3-8B. Memory modules
  trained on MATH, APPS, and MuSiQue generate problem-specific hypotheses
  that guide the solver on the corresponding out-of-distribution benchmarks.
  Results are averaged over three random seeds. CoMem performs best on
  math and code and remains competitive on multi-hop QA.}
  \label{tab:ood}
  \begin{tabularx}{\linewidth}{l*{5}{>{\centering\arraybackslash}X}}
    \toprule
    \rowcolor{tableheadgreen}
    Method & AIME24 & AIME25 & OlympiadBench
      & \shortstack{BigCodeBench-\\Hard}
      & \shortstack{2Wiki\\MultiHopQA} \\
    \midrule
    Base & $16.67^{\pm 0.00}$ & $15.56^{\pm 3.14}$
      & $46.11^{\pm 2.68}$ & $26.35^{\pm 3.76}$ & $56.8^{\pm 0.3}$ \\
    Reflexion & $33.33^{\pm 0.00}$ & $26.67^{\pm 2.72}$
      & $62.50^{\pm 2.50}$ & $39.86^{\pm 2.95}$ & $94.5^{\pm 0.5}$ \\
    DoT-Bank & $33.33^{\pm 2.72}$ & $30.00^{\pm 4.71}$
      & $65.28^{\pm 2.55}$ & $45.50^{\pm 1.03}$ & $92.3^{\pm 1.2}$ \\
    ParamMem & $42.22^{\pm 1.92}$ & $31.11^{\pm 5.09}$
      & $71.11^{\pm 1.27}$ & $45.50^{\pm 0.78}$ & $\mathbf{96.7^{\pm 0.8}}$ \\
    \rowcolor{ragagentgreen}
    CoMem & $\mathbf{44.44^{\pm 3.85}}$ & $\mathbf{36.67^{\pm 0.00}}$
      & $\mathbf{73.06^{\pm 0.96}}$ & $\mathbf{50.23^{\pm 0.39}}$ & $96.1^{\pm 0.5}$ \\
    \bottomrule
  \end{tabularx}
  \centering
  \footnotesize
  \setlength{\tabcolsep}{4pt}
  \renewcommand{\arraystretch}{1.30}
  \caption{Sample efficiency of CoMem. We train the memory
  on 1,000 randomly selected samples.
  Results are averaged over three random seeds. Although performance is
  lower than with full-data training, CoMem outperforms DoT-Bank on
  three benchmarks, suggesting that a small training set can already yield
  useful hypotheses covering a range of
  potential mistakes during problem solving.}
  \label{tab:sample-efficiency}
  \begin{tabularx}{\linewidth}{l*{4}{>{\centering\arraybackslash}X}}
    \toprule
     & \multicolumn{3}{c}{Qwen3-4B} & Qwen3-8B \\
    \cmidrule(lr){2-4}\cmidrule(l){5-5}
    \rowcolor{tableheadgreen}
    Method & MATH & HumanEval+ & LiveCodeBench & MuSiQue \\
    \midrule
    DoT-Bank & $87.77$ & $85.28$ & $50.44$ & $73.5$ \\
    ParamMem & $92.33$ & $92.02$ & $53.53$ & $83.5$ \\
    \rowcolor{ragagentgreen}
    CoMem (full) & $93.05$ & $95.30$
      & $54.67$ & $85.6$ \\
    \rowcolor{ragagentgreen}
    CoMem (1,000 samples) & $91.97$ & $90.80$ & $50.22$ & $79.1$ \\
    \bottomrule
  \end{tabularx}
\end{table}

\cref{tab:ood} shows that CoMem achieves the best performance
on AIME24, AIME25, OlympiadBench, and BigCodeBench-Hard, outperforming
ParamMem and DoT-Bank on all four datasets. On 2WikiMultiHopQA,
CoMem outperforms Reflexion and DoT-Bank and achieves performance
comparable to ParamMem. These results demonstrate that CoMem generalizes
to out-of-distribution tasks using a source-trained memory and references
retrieved from the source-task memory bank.


\obs{Observation 5:} \textbf{Effective guidance from hypotheses with limited supervision.}
We assess whether useful guidance from failure hypotheses can be learned
from small samples. We randomly sample from the questions
that the model answered incorrectly during memory bank construction and
use their supervision targets to form a fine-tuning set of 1,000 examples.
We use the same memory bank as in the full-data setting.

We observe that CoMem retains much of the performance reported in
our main experiments with only 1,000 fine-tuning examples
(\cref{tab:main-results,tab:sample-efficiency}).
Performance remains close on MATH, with larger decreases on the coding
and QA benchmarks. It still outperforms DoT-Bank on MATH, HumanEval+, and
MuSiQue and achieves comparable performance on LiveCodeBench.
This suggests that, even with a limited sample size, the memory can
use references retrieved from similar tasks to produce hypotheses that
cover some of the errors the model
may make when solving test questions. For more challenging problems in
LiveCodeBench, additional hypothesis training examples may be needed to
cover a broader range of potential errors. Results with larger training
sets are provided in \cref{app:sample-efficiency}.


\FloatBarrier

\section{Conclusion}
\label{sec:conclusion}

We introduced CoMem, a self-improvement framework that conditions hypothesis
generation on retrieved experience during training and inference. By
interpreting different retrieved similar cases for the current task, CoMem
provides complementary perspectives on potential errors to support
self-reflection. Experiments across mathematical reasoning, code generation,
and multi-hop QA show strong overall performance and increased hypothesis
diversity. Further evaluations demonstrate out-of-distribution generalization
and weak-to-strong transfer. These findings highlight the value of connecting
cross-task experience with reflection for agent self-improvement. Future work
could incorporate reinforcement learning to jointly improve the correctness
and semantic diversity of generated hypotheses, encouraging coverage of distinct
failure modes while reducing redundant paraphrases of the same error.

\subsection*{Ethics statement}

This work adheres to the ICLR Code of Ethics. Our study uses publicly
available benchmarks for mathematical reasoning, code generation, and
multi-hop question answering and does not involve human-subject or animal
experiments. We report our methods, experimental settings, and results
transparently to support research integrity.

\subsection*{Reproducibility statement}

To facilitate reproducibility, \cref{sec:method} and
\cref{alg:comem-framework} describe the CoMem framework, while
\cref{sec:experimental-setup} and
\cref{app:experimental-details} document the experimental setup,
including dataset sampling, model configurations, training and inference
settings, evaluation procedures, and prompt examples. All benchmarks used
in this study are publicly available. We conduct experiments with three
random seeds and report aggregated results to account for variability.

\bibliography{main}
\bibliographystyle{iclr2027_conference}

\clearpage
\appendix
\section*{Appendix}
\section{Experimental Details}
\label{app:experimental-details}

\subsection{Dataset Details}
\label{app:datasets}

\noindent\textbf{MATH.}
We use MATH~\citep{hendrycks2021math}, a competition-level mathematics
benchmark covering seven subject categories: prealgebra, algebra,
intermediate algebra, geometry, number theory, counting and probability,
and precalculus. We sample up to 1,500 problems per category from the
training split and up to 40 problems per category from the test split.
For out-of-distribution evaluation, we use all problems from AIME24 and
AIME25. For OlympiadBench~\citep{he2024olympiadbench}, we construct a
balanced test subset across four subjects: algebra, combinatorics,
geometry, and number theory. We randomly sample 30 problems per subject,
yielding 120 test problems.


\noindent\textbf{Programming.}
We use the training split of APPS~\citep{hendrycks2021apps} for training
and evaluate on HumanEval+~\citep{liu2023evalplus} and
LiveCodeBench~\citep{jain2025livecodebench}. HumanEval+ contains 164
problems; we exclude HumanEval/32 due to a bug in its test cases, leaving
163 problems for evaluation. For LiveCodeBench, we use release v6 and
sample 300 problems through stratified sampling by platform and
difficulty, allocating samples to each stratum in proportion to its size
in the original release.

\noindent\textbf{Multi-hop QA.}
We adopt HotpotQA~\citep{yang2018hotpotqa} and
MuSiQue~\citep{trivedi2022musique}. For HotpotQA, we use 30,000 questions
from the distractor training split for training. We construct a held-out
test set from the remaining questions by sampling 100 questions from each
of the easy, medium, and hard difficulty levels, yielding 300 test questions.
For MuSiQue, we sample 10,000 training questions across different reasoning
types. The test set contains 180 questions, with 60 each for 2-hop, 3-hop,
and 4-hop reasoning.
For out-of-distribution evaluation, we construct a balanced evaluation set
of 200 questions from 2WikiMultiHopQA~\citep{ho2020twikimultihopqa},
with 50 questions from each of four reasoning types:
bridge-comparison, comparison, compositional, and inference.


\subsection{Implementation Details}
\label{app:implementation}

We run all experiments with three random seeds: 42, 43, and 44.
During evaluation, we use a solving budget of five attempts per stage,
including the initial attempt.
For CoMem, we set $k=5$ during both training and inference and $V=8$
at inference. We use a retrieval softmax temperature of
$T_{\mathrm{ret}}=0.1$ during both training and inference.
During evaluation, we set
the sampling temperature for task solving to
$T = 0.2$ in the first iteration and $T = 1.0$ in subsequent iterations
to promote diversity. The maximum generation length is 2,048 tokens for
MATH and HumanEval+, 4,096 tokens for LiveCodeBench and BigCodeBench-Hard,
8,192 tokens for AIME24 and AIME25, and 32,000 tokens for HotpotQA and MuSiQue.


\noindent\textbf{Memory bank construction.}
We construct a memory bank for each domain and backbone model using the
corresponding training data. Problems solved on the first attempt are
excluded. If the initial attempt fails, the model generates reflections
based on the feedback and attempts to solve the problem again. We allow
up to $K=3$ attempts in total, including the initial attempt.
Successful attempts are stored as positive entries, while final
unsuccessful attempts are stored as negative entries. Each entry contains
a failed solution, its corresponding reflection, and the updated solution.
For MATH, the Qwen3-4B bank contains 2,809 entries (1,161 positive and
1,648 negative), and the Qwen3-8B bank contains 2,733 entries (1,359
positive and 1,374 negative). For programming, the corresponding banks
contain 3,264 entries (953 positive and 2,311 negative) and 3,349 entries
(1,361 positive and 1,988 negative), respectively.

\subsubsection{Hypothesis Supervision Prompt}
\label{app:pitfall-label-prompt}

\begin{promptbox}[ignore nobreak,before={},label={box:pitfall-label-prompt-math}]{Hypothesis Supervision Prompt (MATH)}
You are a pitfalls predictor. Given a math problem, list up to 6 potential
common reasoning mistakes someone might make on the CURRENT problem.
Do not solve the current problem, do not compute or reveal the final answer,
and do not include solution steps. Generate only pitfalls. Each item must
be specific to the current problem, not generic study advice. Do not list
correct facts, valid reasoning steps, or true intermediate observations as
mistakes. Avoid self-contradictory items. You may receive retrieved solved
and unsolved examples as optional reference material. They can help suggest
possible reasoning risks for the current problem, though their relevance
may vary. Base the final pitfalls primarily on the current problem, and
adapt useful signals from the examples rather than mechanically copying
them. Output ONLY a Markdown numbered list, items 1-6, with no preamble,
closing remarks, wrapper tags, or restatement of the problem. An earlier
problem-and-pitfalls exchange is provided as a one-shot demonstration.
It may guide the output format and level of specificity, while the pitfalls
should be tailored to the current problem and its retrieved references.
\end{promptbox}

\begin{promptbox}[label={box:pitfall-label-prompt-code}]{Hypothesis Supervision Prompt (Code)}
You are a pitfall predictor for Python coding tasks. You are given a CURRENT
coding task plus two retrieved reference tasks: one similar SOLVED task with a
successful implementation and reflection, and one similar UNSOLVED task with a
failed implementation and reflection. The retrieved tasks are provided
specifically as references for predicting pitfalls in the CURRENT task.
Actively transfer relevant implementation patterns from the solved reference
and relevant failure modes from the unsolved reference. Do not blindly copy
them: account for differences in the task statement, interface, constraints,
and input/output contract, and ignore irrelevant details. Every item must be
specific to the CURRENT task. Return exactly 4 to 6 concise, non-overlapping
pitfalls under exactly one \texttt{<Pitfalls>:} heading. Return pitfalls only.
Do not output code, an implementation, a complete algorithm, a proof, or a
final answer.
\end{promptbox}

\begin{promptbox}[label={box:pitfall-label-prompt-qa}]{Hypothesis Supervision Prompt (QA)}
You are a pitfalls predictor for multi-hop factual QA. Each request includes
one similar solved reflection transition and one similar unsolved reflection
transition retrieved by question similarity. Treat the solved transition as
a validated example of an error that was corrected. Treat the unsolved
transition as warning evidence that may be incomplete or mistaken; do not
blindly imitate it. Use both only to identify transferable risks for the
CURRENT question. The CURRENT question includes its evidence context
paragraphs (gold and distractor). Ground each pitfall in the actual
entities and passages present in the context where possible. When referring
to entities from the context, use paragraph indices or descriptive
references (e.g., ``the entity in [3]''); never write out any string that
could be the final answer. List 4 to 6 plausible reasoning or retrieval
pitfalls that are specific to the CURRENT question. Focus on misreading
constraints, confusing entities or relations, skipping a reasoning hop,
using insufficient evidence, and returning the wrong answer type. Do NOT
solve the CURRENT question, state or guess a candidate answer, reveal a
final conclusion, or provide a step-by-step solution. Output ONLY a
Markdown numbered list of 4--6 consecutive items starting at 1, with no
preamble, headings, closing remarks, or wrapper tags.
\end{promptbox}

\begin{promptbox}[label={box:math-oneshot-example}]{One-shot Demonstration (MATH)}
\textbf{User}

\textbf{=== Similar solved example ===}

Problem:\\
If $t(x) = 9+2f(x)$ and $f(x) = x^2-3$, then what is $t(f(-2))$?

Successful solution sketch:\\
Evaluates the nested composition from the inside out, then applies the
definition of $t$ without conflating $f(-2)$ with the later evaluation of $f$.

Prior reflection / key principle:\\
Function-composition inputs must be propagated one layer at a time;
substituting into the outer expression too early changes the computation.

\textbf{=== Similar unsolved example ===}

Problem:\\
A piecewise graph of $f$ on $[-4,4]$ is given. Count the values of $x$
satisfying $f(f(x))=2$.

Historical failed attempt:\\
Located several inputs of the outer function but did not consistently solve
every corresponding inner preimage across all pieces of the graph.

Observed pitfall / reflection:\\
For a composition equation, first enumerate every value mapped to the target
by the outer function, then find all valid preimages of each value.

\textbf{=== One-shot demonstration problem ===}

If $f(x) = 8x^3 - 6x^2 - 4x + 5$, find the value of $f(-2)$.

List up to 6 potential pitfalls / common mistakes for this problem. Do not solve it.
\tcblower
\textbf{Assistant}

\begin{enumerate}[leftmargin=1.6em,itemsep=3pt,parsep=0pt,topsep=3pt]
  \item \textbf{Substituting the wrong input} - Evaluating \texttt{f(2)} or
  another nearby value instead of substituting \texttt{x = -2}.
  \item \textbf{Mishandling powers of a negative number} - Computing
  \texttt{(-2)\textasciicircum{}3} or \texttt{(-2)\textasciicircum{}2} with the wrong sign.
  \item \textbf{Losing the coefficient signs} - Treating
  \texttt{-6(-2)\textasciicircum{}2} or \texttt{-4(-2)} as if the coefficient signs do not matter.
  \item \textbf{Violating order of operations} - Adding terms before
  evaluating powers and products.
  \item \textbf{Omitting a polynomial term} - Forgetting to include
  \texttt{-4x} or the constant \texttt{+5} in the evaluation.
  \item \textbf{Combining terms arithmetically incorrectly} - Making small
  addition or subtraction errors when summing the terms.
\end{enumerate}
\end{promptbox}

\subsubsection{Memory Training Format}
\label{app:pitfall-training-format}

\begin{promptbox}[label={box:pitfall-training-format-math}]{Memory Training Format (MATH)}
You are a pitfalls predictor. Given a math problem, list up to 6 potential
common reasoning mistakes someone might make on the CURRENT problem.
Do not solve the current problem, do not compute or reveal the final answer,
and do not include solution steps. Generate only pitfalls. Each item must
be specific to the current problem, not generic study advice. Do not list
correct facts, valid reasoning steps, or true intermediate observations as
mistakes. Avoid self-contradictory items. You may receive retrieved solved
and unsolved examples as optional reference material. They can help suggest
possible reasoning risks for the current problem, though their relevance
may vary. Base the final pitfalls primarily on the current problem, and
adapt useful signals from the examples rather than mechanically copying
them. Output ONLY a Markdown numbered list, items 1-6, with no preamble,
closing remarks, wrapper tags, or restatement of the problem.
\end{promptbox}

\begin{promptbox}[label={box:pitfall-training-format-code}]{Memory Training Format (Code)}
You are a proactive coding-task analyst.

Given one current coding task and two retrieved reference tasks, predict a
compact set of task-specific pitfalls that would help a solver avoid likely
mistakes before writing the solution.

Requirements:
\begin{itemize}
  \setlength{\itemsep}{0pt}
  \setlength{\parsep}{0pt}
  \item[-] Return only a \texttt{<Pitfalls>:} section.
  \item[-] Write 1 to 6 numbered pitfalls. Use fewer items when the available
  information does not support more.
  \item[-] Make each item concrete and actionable for the current task.
  \item[-] Cover distinct failure modes instead of paraphrasing one issue.
  \item[-] Do not write code or solve the task.
\end{itemize}
\end{promptbox}

\begin{promptbox}[label={box:pitfall-training-format-qa}]{Memory Training Format (QA)}
You are an AI assistant specialized in multi-hop factual QA. Given a
current question with its evidence context paragraphs and retrieved
similar solved/unsolved reflection transitions, identify and explain
4 to 6 reasoning or retrieval pitfalls that a solver might encounter
on the current question. Ground pitfalls in specific entities and
passages from the context where possible. A solved transition shows a
validated correction; an unsolved transition is warning evidence that
may be incomplete. Do not solve the current question, state or guess
a candidate answer, reveal a conclusion, or provide solution steps.
Output only a Markdown numbered list.
\end{promptbox}

\subsubsection{QA Semantic Judge Prompt}
\label{app:qa-judge-prompt}

\begin{promptbox}[label={box:qa-semantic-prompt}]{QA Semantic Judge Prompt}
Determine if the following two answers are semantically similar or equivalent
in the context of a factual QA task.
Let's not be too strict. For instance:
\begin{enumerate}
  \setlength{\itemsep}{0pt}
  \setlength{\parsep}{0pt}
  \item ``Aroostook'' and ``Aroostook County, Maine, United States'' should be
  considered similar;
  \item Spinning industry and spining should be considered similar.
  \item Some abbreviations and their full forms should be considered
  equivalent. For instance, ``Britain'', ``UK'', and ``United Kingdom''
  are similar.
  \item Some fine-grained categories and their corresponding coarse-grained
  categories should be considered similar. e.g., ``musician'' and ``musical
  artist'' are similar. ``music'' and ``rock music'' are similar.
\end{enumerate}
If the two answers are semantically similar or refer to the same entity/fact,
judge them as equivalent and respond ``Yes''. Only respond ``No'' when they
clearly refer to different things.

Answer 1: \texttt{\{prediction\}}\\
Answer 2: \texttt{\{gold\}}

Respond only with ``Yes'' or ``No''.
\end{promptbox}

\subsection{Framework Pseudocode}
\label{app:framework-pseudocode}

\cref{alg:comem-framework} summarizes CoMem following
\cref{sec:method}. The frozen base model $p_\theta$ collects
experience, generates supervision, and serves as the solver. Only the
LoRA adapter of the memory $G_\phi$ is trained.
The training bank $\mathcal{B}$ stays fixed during evaluation, while
$\mathcal{C}$ accumulates successful test-time trajectories.

\begin{algorithm}[!htbp]
  \caption{CoMem Framework}
  \label{alg:comem-framework}
  \small
  \begin{algorithmic}[1]
    \Require Training tasks $\mathcal{D}_{\mathrm{tr}}$, supervision tasks
    $\mathcal{D}_{\mathrm{sup}} \subseteq \mathcal{D}_{\mathrm{tr}}$,
    test tasks $\mathcal{D}_{\mathrm{te}}$
    \Require Base model $p_\theta$, bank budget $K=3$, per-stage solving budget $K_s=5$,
    shortlist size $k=5$, number of hypothesis sets $V=8$,
    retrieval softmax temperature $T_{\mathrm{ret}}=0.1$
    \Ensure Memory $G_\phi$, training bank $\mathcal{B}$,
    results $\mathcal{Y}$
    \Statex \textbf{Memory bank construction}
    \State $\mathcal{B}^{+} \gets \emptyset$, $\mathcal{B}^{-} \gets \emptyset$
    \For{$x \in \mathcal{D}_{\mathrm{tr}}$}
      \State $(y,f,s) \gets \Call{SolveAndEvaluate}{p_\theta,x}$
      \If{$s=-$} \Comment{Skip initially solved tasks}
        \State $(\tau,s) \gets \Call{ReflectAndRetry}{p_\theta,x,y,f,K-1}$
        \State $\mathcal{B}^{s} \gets \mathcal{B}^{s} \cup \{(x,\tau)\}$
      \EndIf
    \EndFor
    \State $\mathcal{B} \gets \mathcal{B}^{+} \cup \mathcal{B}^{-}$
    \Statex \textbf{Supervision and training}
    \State $\mathcal{S} \gets \emptyset$
    \For{$x \in \mathcal{D}_{\mathrm{sup}}$}
      \State $\mathcal{L}_x^{\pm} \gets \Call{TopK}{\mathcal{B}^{\pm},x,k}$ \Comment{Exclude $x$ itself}
      \State $p_x^{\pm} \gets \operatorname{softmax}(\operatorname{sim}(x,\mathcal{L}_x^{\pm})/T_{\mathrm{ret}})$
      \State $R_x \gets \Call{SamplePair}{p_x^{+},p_x^{-}}$
      \State $z_x \sim p_\theta(\cdot \mid x,R_x)$
      \State $\mathcal{S} \gets \mathcal{S} \cup \{((x,R_x),z_x)\}$
    \EndFor
    \State Discard generic advice from the generated supervision $\mathcal{S}$.
    \State $G_\phi \gets \Call{LoRATrain}{p_\theta,\mathcal{S}}$ \Comment{\cref{eq:sft}}
    \Statex \textbf{Inference: retrieval and hypothesis generation}
    \For{$x \in \mathcal{D}_{\mathrm{te}}$}
      \State $\mathcal{L}_x^{\pm} \gets \Call{TopK}{\mathcal{B}^{\pm},x,k}$
      \State $p_x^{\pm} \gets \operatorname{softmax}(\operatorname{sim}(x,\mathcal{L}_x^{\pm})/T_{\mathrm{ret}})$
      \For{$v=1,\ldots,V$} \Comment{Independent draws with replacement}
        \State $R_x^{(v)} \gets \Call{SamplePair}{p_x^{+},p_x^{-}}$
        \State $h_x^{(v)} \sim G_\phi(\cdot \mid x,R_x^{(v)})$
      \EndFor
      \State $\mathcal{H}_x \gets (h_x^{(1)},\ldots,h_x^{(V)})$
    \EndFor
    \State $\mathcal{C} \gets \emptyset$, $\mathcal{F} \gets \emptyset$, $\mathcal{Y} \gets \emptyset$
    \Statex \textbf{Stage 1: Hypothesis-guided solving}
    \For{$x \in \mathcal{D}_{\mathrm{te}}$}
      \State $r_{1:0} \gets ()$ \Comment{No reflection before the first attempt}
      \For{$t=1,\ldots,K_s$}
        \State $y_t \sim p_\theta(\cdot \mid x,r_{1:t-1},h_x^{(t)})$
        \State $(f_t,s) \gets \Call{Evaluate}{x,y_t}$
        \If{$s=+$}
          \State $\tau \gets (y_{1:t},r_{1:t-1})$
          \State $\mathcal{C} \gets \mathcal{C} \cup \{(x,\tau)\}$
          \State \textbf{break}
        \EndIf
        \State \textbf{if} $t<K_s$ \textbf{then} $r_t \gets \Call{Reflect}{p_\theta,x,y_t,f_t}$
      \EndFor
      \State $\mathcal{Y}[x] \gets (y_t,s)$
      \State \textbf{if} $s=-$ \textbf{then} $\mathcal{F} \gets \mathcal{F} \cup \{x\}$
    \EndFor
    \Statex \textbf{Stage 2: Retry with successful trajectories}
    \For{$x \in \mathcal{F}$} \Comment{After all Stage 1 tasks finish}
      \State \textbf{if} $\mathcal{C}=\emptyset$ \textbf{then break}
      \State $\tau^* \gets \Call{RetrieveMostSimilar}{\mathcal{C},x}$
      \State $j \sim \operatorname{Uniform}(\{1,\ldots,V\})$
      \State $(y,\tau,s) \gets \Call{RetryWithTrajectory}{p_\theta,x,h_x^{(j)},\tau^*,K_s}$
      \State $\mathcal{Y}[x] \gets (y,s)$
      \State \textbf{if} $s=+$ \textbf{then} $\mathcal{C} \gets \mathcal{C} \cup \{(x,\tau)\}$
    \EndFor
    \State \Return $(G_\phi,\mathcal{B},\mathcal{Y})$
  \end{algorithmic}
\end{algorithm}

Here $s \in \{+,-\}$ indicates correctness and $f$ is the task feedback.
The budgets $K$ and $K_s$ include the initial attempt, and $K_s$ applies
separately to each solving stage.
\textsc{ReflectAndRetry} makes at most $K-1$ further attempts and returns
the successful repair transition or the final failed attempt and its
reflection, as described in \cref{sec:building}.

For supervision generation, \textsc{TopK} retrieves the top-$k$ tasks separately
from $\mathcal{B}^{+}$ and $\mathcal{B}^{-}$, excluding $x$ itself.
We divide the cosine similarity scores by $T_{\mathrm{ret}}$ before
applying softmax within each shortlist.
\textsc{SamplePair} independently draws one positive and one negative
entry according to their respective probabilities to form $R_x$.
The references remain in the LoRA input. At inference, we again retrieve
top-$k$ shortlists and compute softmax probabilities.
We repeat \textsc{SamplePair} independently $V$ times with replacement,
so reference pairs may repeat. Each sampled pair conditions one hypothesis
set, and these sets form the ordered tuple $\mathcal{H}_x$.
During the reflection loop, each attempt uses the next set and all
preceding reflections.

\Needspace{5\baselineskip}
\textsc{RetryWithTrajectory} denotes one additional retry procedure from
\cref{sec:solving}, using the retrieved successful trajectory and
one hypothesis set sampled uniformly from $\mathcal{H}_x$.
It is called once per unresolved task when a successful reference is
available and allows at most $K_s$ solving attempts. Newly solved trajectories enter
$\mathcal{C}$ immediately for subsequent retries.

\section{Additional Results}
\label{app:additional-results}

\subsection{Additional Diversity Results}
\label{app:diversity}

\subsubsection{Distance-Based Diversity Analysis}
\label{app:diversity-distances}

\begin{figure}[!tbp]
  \centering
  \begin{minipage}[t]{0.48\linewidth}
    \centering
    \includegraphics[width=\linewidth]{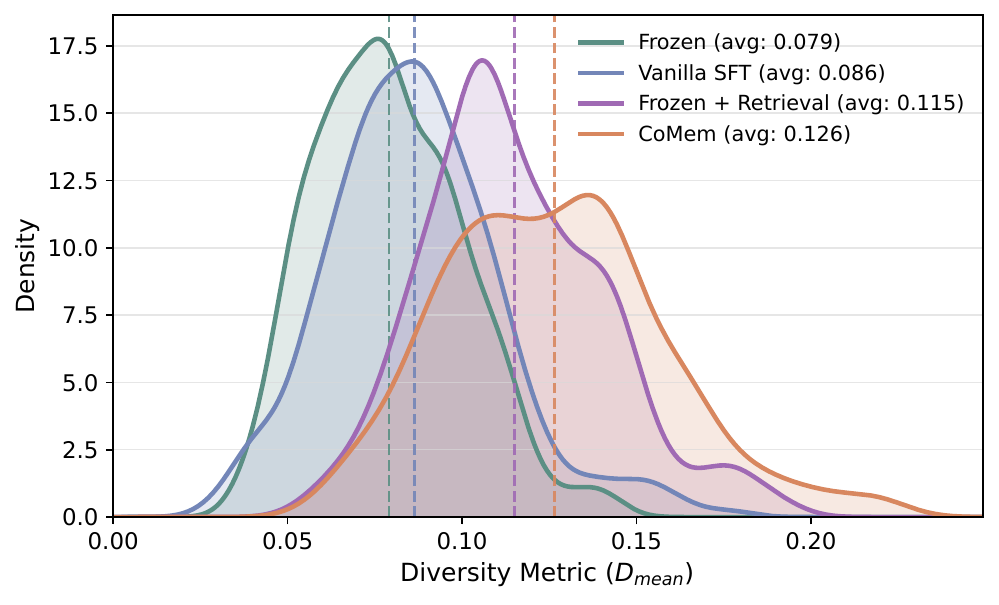}
    \par\small (a) MATH -- Qwen3-4B
  \end{minipage}\hfill
  \begin{minipage}[t]{0.48\linewidth}
    \centering
    \includegraphics[width=\linewidth]{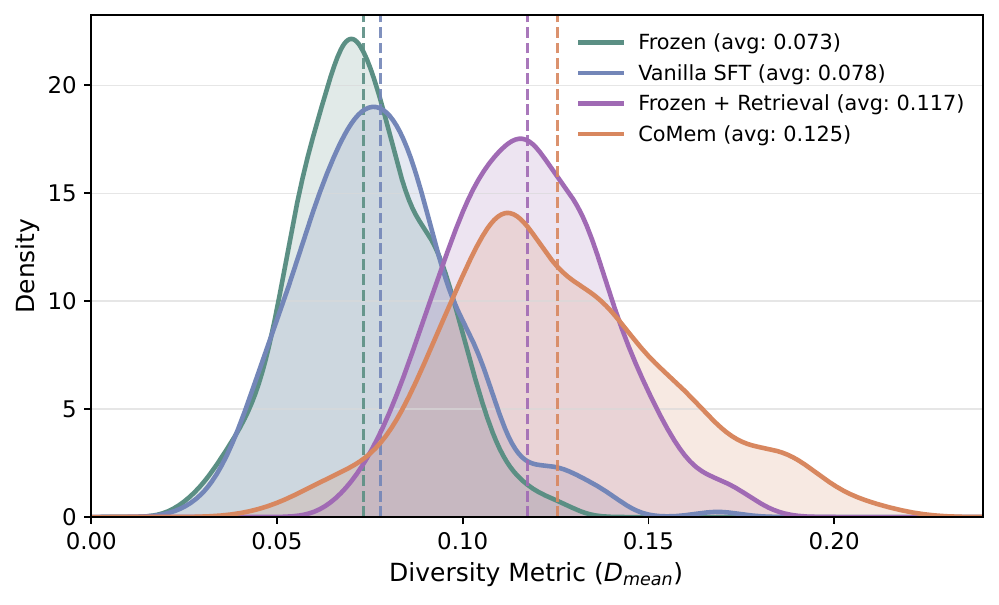}
    \par\small (b) MATH -- Qwen3-8B
  \end{minipage}
  \par\medskip
  \begin{minipage}[t]{0.48\linewidth}
    \centering
    \includegraphics[width=\linewidth]{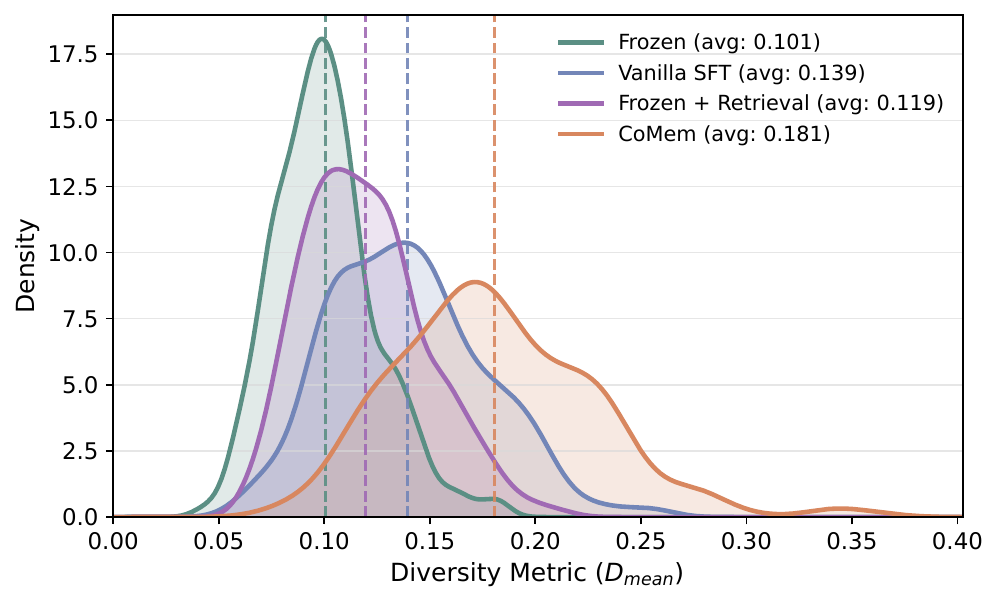}
    \par\small (c) LiveCodeBench -- Qwen3-4B
  \end{minipage}\hfill
  \begin{minipage}[t]{0.48\linewidth}
    \centering
    \includegraphics[width=\linewidth]{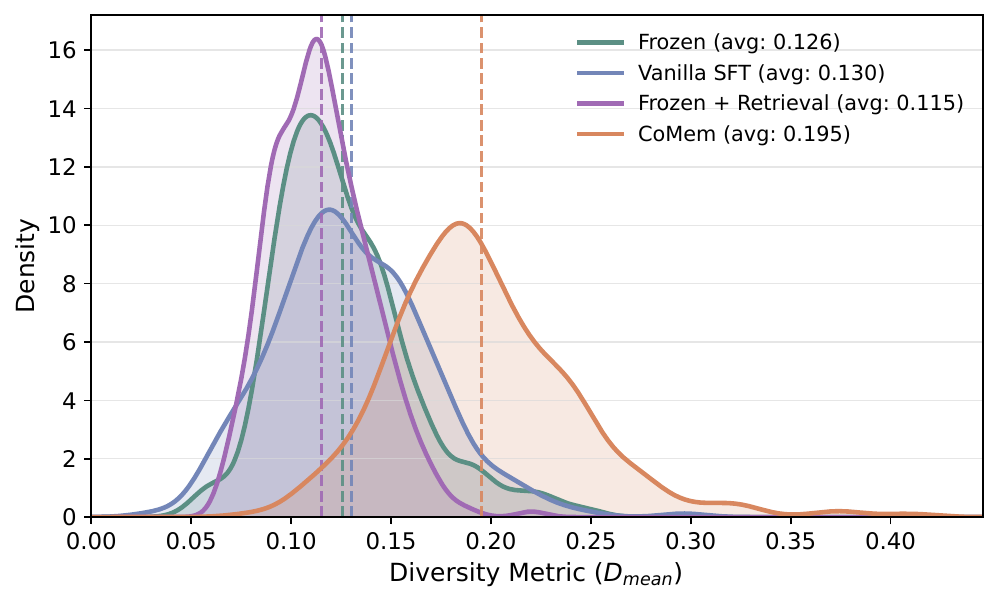}
    \par\small (d) LiveCodeBench -- Qwen3-8B
  \end{minipage}
  \par\medskip
  \begin{minipage}[t]{0.48\linewidth}
    \centering
    \includegraphics[width=\linewidth]{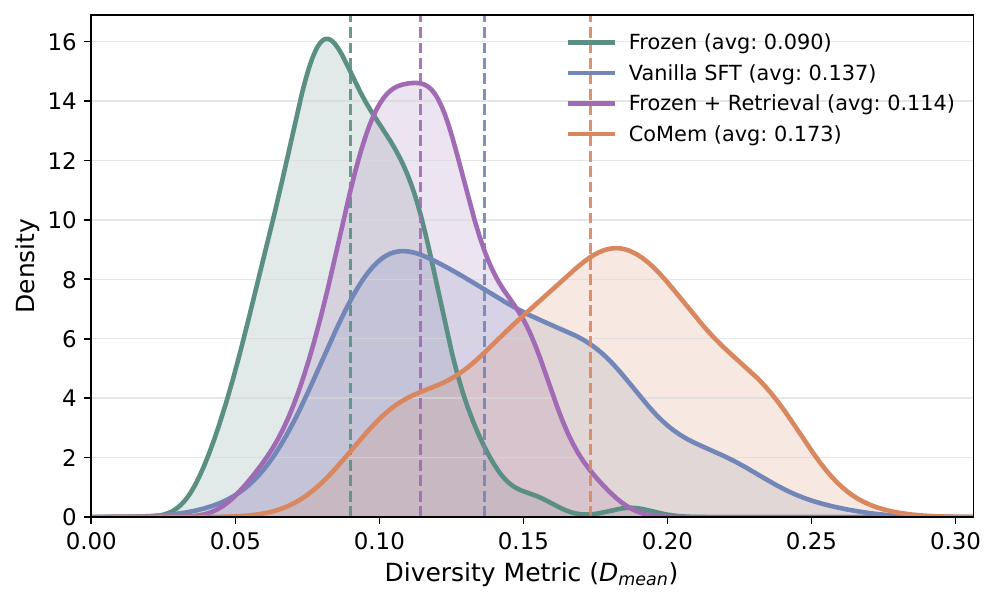}
    \par\small (e) HumanEval+ -- Qwen3-4B
  \end{minipage}
  \caption{Additional hypothesis diversity results on MATH and code generation.
  The top row shows MATH with Qwen3-4B and Qwen3-8B, the middle row shows
  LiveCodeBench with the same backbones, and the bottom row shows HumanEval+
  with Qwen3-4B. Dashed lines mark mean $D_{\mathrm{mean}}$ scores.
  All configurations generate 10 hypothesis sets per question at $T=1.0$.}
  \label{fig:diversity-additional}
\par
\begingroup
  \centering
  \footnotesize
  \setlength{\tabcolsep}{4pt}
  \renewcommand{\arraystretch}{1.10}
  \captionof{table}{CoMem performance across fine-tuning dataset sizes with
  the memory bank held fixed. Values are final accuracy for MATH and
  MuSiQue, and Pass@1 for HumanEval+ and LiveCodeBench, in percentages.
  Full denotes training on the full dataset. Dashes indicate unavailable
  results. Bold marks the best result within each dataset.}
  \label{tab:sample-efficiency-additional}
  \begin{center}
  \begin{tabularx}{\linewidth}{@{}l*{4}{>{\centering\arraybackslash}X}@{}}
    \toprule
     & \multicolumn{3}{c}{Qwen3-4B} & Qwen3-8B \\
    \cmidrule(lr){2-4}\cmidrule(l){5-5}
    \rowcolor{tableheadgreen}
    Training samples & MATH & HumanEval+ & LiveCodeBench & MuSiQue \\
    \midrule
    2,000 & $\mathbf{93.65}$ & $92.02$ & $51.78$ & -- \\
    3,000 & $92.45$ & -- & -- & $80.7$ \\
    Full & $93.05$ & $\mathbf{95.30}$ & $\mathbf{54.67}$ & $\mathbf{85.6}$ \\
    \bottomrule
  \end{tabularx}
  \end{center}
\endgroup
\end{figure}

\cref{fig:diversity-additional} shows that CoMem achieves the highest
mean hypothesis diversity in all five settings. On MATH, Frozen + Retrieval
exceeds both Frozen and Vanilla SFT with both backbones, with larger gains
over Frozen than on the coding benchmarks. These results suggest that
retrieved mathematical experience provides a stronger source of variation
in hypothesis generation. CoMem further increases diversity as its memory
learns to relate this experience to the current task and derive different
failure hypotheses from it.

For each question $x$, let $\mathbf{e}_i$ denote the embedding of its
$i$-th generated hypothesis set. With $K=10$ sets per question, we compute
\begin{equation}
  D_{\mathrm{mean}}(x) = \frac{2}{K(K-1)}
  \sum_{1 \leq i < j \leq K}
  \left(1 - \frac{\mathbf{e}_i^{\top}\mathbf{e}_j}
  {\lVert\mathbf{e}_i\rVert_2\lVert\mathbf{e}_j\rVert_2}\right).
\end{equation}
The plots show the distribution of these question-level scores.


\subsubsection{Clustering-Based Diversity Analysis}
\label{app:diversity-clustering}

To complement the pairwise-distance analysis, we apply k-means to the
embeddings of generated hypothesis sets. For each problem, we jointly cluster
40 sets, comprising 10 sets from each of Frozen, Vanilla SFT,
Frozen + Retrieval, and CoMem, with $K \in \{2,3,5,8,10\}$. This shared partition allows the
methods to be compared at the same clustering granularity. We compute
the entropy-based effective cluster count for each method within this
partition and average it across problems.

\begin{figure}[!t]
  \centering
  \includegraphics[width=\linewidth]{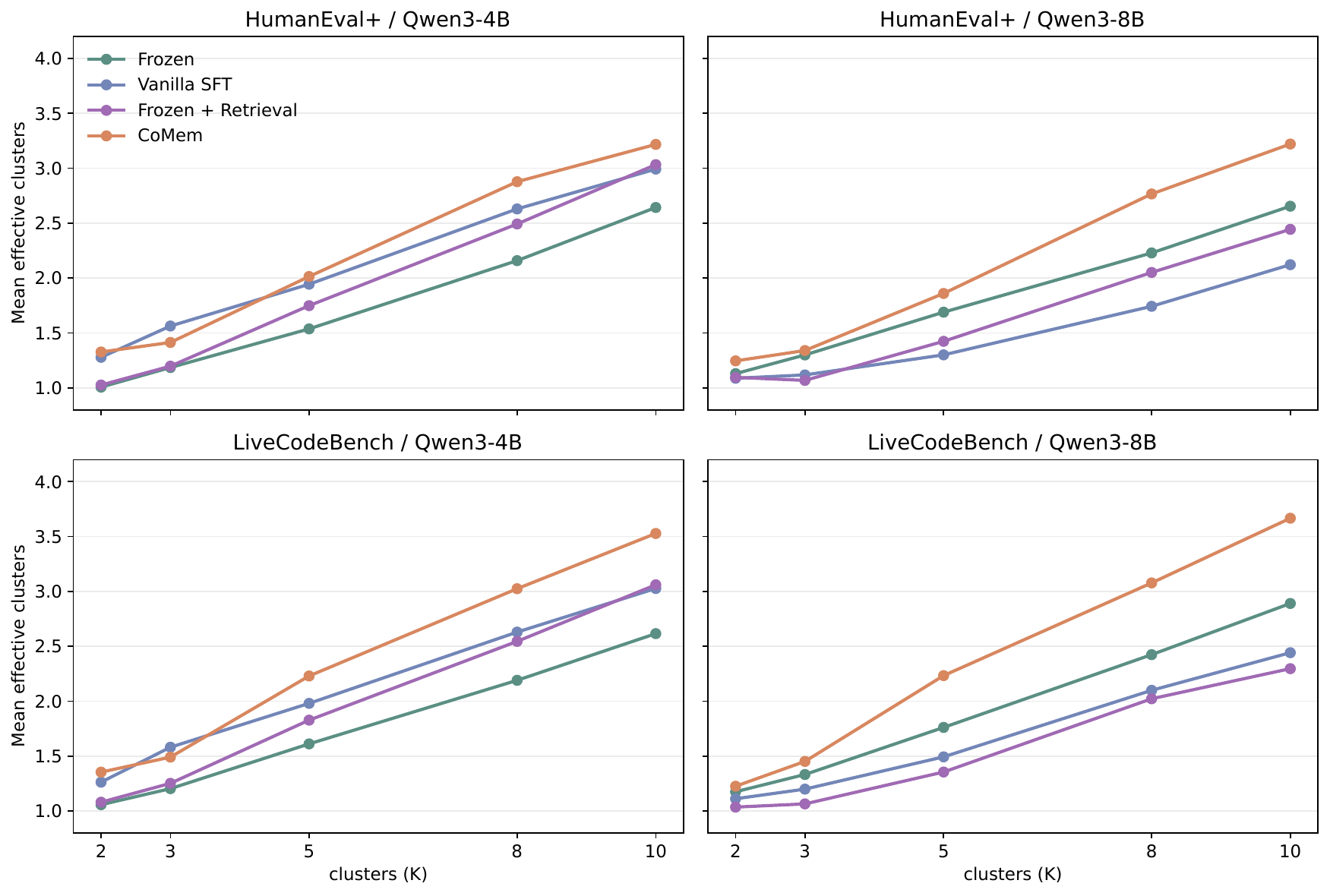}
  \caption{Clustering-based diversity of generated hypothesis sets on HumanEval+
  and LiveCodeBench with Qwen3-4B and Qwen3-8B. Curves show the mean
  entropy-based effective cluster count across problems. For each problem,
  k-means is applied jointly to 40 hypothesis sets, with 10 sets from each
  of Frozen, Vanilla SFT, Frozen + Retrieval, and CoMem, using $K$ shared clusters.}
  \label{fig:diversity-clustering}
\end{figure}

As shown in \cref{fig:diversity-clustering}, at $K=5,8,$ and $10$,
CoMem achieves the highest mean effective cluster counts in all four
settings. With Qwen3-4B, Vanilla SFT is slightly higher at $K=3$.
With Qwen3-8B, Frozen achieves higher effective cluster counts than
Vanilla SFT and Frozen + Retrieval, while CoMem remains highest across
all tested values of $K$. These results complement the distance-based
analysis, showing that CoMem's hypothesis sets are less concentrated
in a few clusters.

\subsection{Additional Sample-Efficiency Results}
\label{app:sample-efficiency}

The source training set for programming contains only 5,000 problems
and is smaller than those for MATH and QA. We therefore evaluate an
additional 2,000-sample setting for HumanEval+ and LiveCodeBench, while
considering both 2,000 and 3,000 samples for MATH and 3,000 samples for
MuSiQue. We use Qwen3-4B for MATH and programming, and Qwen3-8B for
MuSiQue.

We observe that using 2,000 training samples on MATH yields slightly
better performance than full-data training
(\cref{tab:sample-efficiency-additional}), and the 3,000-sample
setting achieves performance close to the full-data result.
A small training set, combined with retrieved references, may already
provide useful coverage of common errors on MATH.
In contrast, on HumanEval+ and LiveCodeBench, using 2,000 samples improves
performance over the 1,000-sample setting but still
falls short of full-data training. For more challenging programming
problems, limited training data may leave some failure modes
insufficiently represented, restricting the memory's ability to
generate task-specific failure hypotheses. Additional training examples could
broaden this coverage and help the memory identify potential mistakes
across different programming problems. On MuSiQue, the larger full
training set may cover a broader range of potential errors than the
3,000-sample subset. Although training on this subset falls short of
full-data performance, it still outperforms DoT-Bank, indicating that
limited supervision can yield useful hypotheses.


\subsection{Case Study}
\label{app:case-study}

We use case studies to illustrate how CoMem promotes diversity in both
hypotheses and reflections. By highlighting different potential errors,
hypotheses can encourage reflections to consider alternative explanations
and overlooked requirements, guiding solution revision.
We present hypotheses and reflections alongside the resulting
answers or code.

\subsubsection{MATH}
\label{app:case-math}

We show two tasks in MATH where CoMem revises incorrect answers after
reflection. These cases illustrate checks on the symmetry of an equation
and constraints over an entire interval,
with comparisons against Reflexion, DoT-Bank, and ParamMem.


\Needspace{6\baselineskip}
\subsubsubsection{\textbf{Case 1: Revisiting Symmetry in a Rational Equation}}
\label{app:case-math-symmetry}

\textbf{Task.} Find the sum of all complex solutions to
\[
  \sum_{k=1}^{4}\frac{k}{x^2-k}=2010x-4.
\]
The correct sum is $4/2010=2/1005$.

\begin{casepanel}{Case 1: Baselines}
\textbf{Reflexion.} A reflection recognizes the broken symmetry. The
final solution nevertheless misclassifies the equation as transcendental
and returns 0 without carrying out the polynomial reduction.

\textbf{DoT-Bank.} A reflection identifies the linear term $2010x$ as
breaking symmetry. The final solution misidentifies the polynomial's
degree after clearing denominators and returns 0 without computing the
required coefficients.

\textbf{ParamMem.} Reflections also identify the asymmetry. The final
solution acknowledges that $x$ and $-x$ need not be paired, but appeals
to an unsupported cancellation argument and returns 0.
\end{casepanel}

\Needspace{6\baselineskip}
\begin{casepanel}[casesuccess]{Case 1: CoMem}
\textbf{Hypotheses.} The guidance highlights sign and coefficient errors
when handling the right-hand side $2010x-4$, and warns against applying
Vieta's formula to an incorrectly formed polynomial.

\textbf{Reflection.} ``Failing to recognize that the right-hand side
introduces an odd-degree term, disrupting the expected symmetry.''
\tcblower
\textbf{Solution revision.} The preceding solution treats the equation
as containing only even powers and returns 0. The revision rejects this
symmetry assumption and changes the answer to $4/2010$.

\textbf{Result.} The revised answer matches the reference.
\end{casepanel}

The hypotheses flag risks in forming the polynomial, and the reflection
identifies the symmetry error. The baselines also recognize this issue
but retain an incorrect final answer, while CoMem's revision corrects it.

\Needspace{6\baselineskip}
\subsubsubsection{\textbf{Case 2: Checking a Polynomial Across the Entire Interval}}
\label{app:case-math-polynomial}

\textbf{Task.} For real $a,b,c$, suppose
$|ax^2+bx+c|\leq1$ for every $x\in[0,1]$. Find the largest possible
value of $|a|+|b|+|c|$. The correct answer is 17.

\begin{casepanel}{Case 2: Baselines}
\textbf{Reflexion.} A reflection calls for checking the entire interval.
The revised solution transforms the degree-two Chebyshev polynomial to
$[0,1]$ and returns 17.

\textbf{DoT-Bank.} Reflections also raise the need to check interior
points. The final solution nevertheless uses $P(x)=-4x^2+2x+1$ and
returns 7, claiming that its checks at $0$, $1/2$, and $1$ ensure
the bound everywhere. This candidate violates the constraint, since
$P(1/4)=5/4>1$.

\textbf{ParamMem.} The final solution constructs $8x^2-8x+1$ and
returns 17. No reflection is recorded.
\end{casepanel}

\begin{casepanel}[casesuccess]{Case 2: CoMem}
\textbf{Hypotheses.} The guidance covers interior extrema, the
applicability of Chebyshev polynomials on $[0,1]$, dependencies between
coefficients, and the need to consider both upper and lower bounds.

\textbf{Reflection.} ``Assuming the maximum occurs only at the endpoints
$x=0$ and $x=1$, neglecting the behavior of the polynomial at intermediate
points like $x=0.5$ or critical points within the interval.''
\tcblower
\textbf{Solution revision.} The preceding answer is 8. The revision
considers the whole interval and maps $T_2(t)=2t^2-1$ from $[-1,1]$
to $[0,1]$, giving
\[
  P(x)=T_2(2x-1)=8x^2-8x+1.
\]
This polynomial satisfies the bound throughout $[0,1]$, and its
coefficient magnitudes sum to $8+8+1=17$.

\textbf{Result.} The revised answer is 17, matching the reference.
\end{casepanel}

The hypotheses suggest several ways to check the constrained polynomial.
The reflection emphasizes the interior of the interval, and
the revised answer constructs a polynomial satisfying the full constraint.
DoT-Bank raises a similar check but does not enforce it in its final solution.

\subsubsection{HumanEval+}
\label{app:case-humaneval}

We show two tasks in HumanEval+ that illustrate checks on signed-digit
semantics and duplicate removal. CoMem corrects faulty code in both cases,
while Reflexion, DoT-Bank, and ParamMem remain unsuccessful.


\subsubsubsection{\textbf{Case 3: Handling the Sign of the First Digit}}
\label{app:case-signed-digits}

\textbf{Task.} \texttt{count\_nums} counts integers whose signed-digit sum
is positive. For a negative integer, only the first digit is negative.
Thus, $-123$ contributes $-1+2+3=4$ and is counted, whereas $-11$
contributes $-1+1=0$ and is excluded.

\Needspace{6\baselineskip}
\begin{casepanel}{Case 3: Baselines}
\textbf{Reflexion.} The final program applies the number's sign to the
entire digit sum, incorrectly excluding $-123$.

\textbf{DoT-Bank.} The final program also negates the sum of all digits
for a negative integer, retaining the same sign-handling error.

\textbf{ParamMem.} A reflection recommends: ``When processing a negative
number, extract the first signed digit correctly by treating it as
negative and then summing the absolute values of the remaining digits.''
The final program nevertheless subtracts 1 and adds all unsigned digits,
incorrectly counting $-11$.
\end{casepanel}

\begin{casepanel}[casesuccess]{Case 3: CoMem}
\textbf{Hypotheses.} The saved hypotheses cover signed-digit contributions,
distinguishing positive from nonzero sums, and preserving the integer
return type.

\textbf{Reflection.} ``Adjust the logic for extracting the first digit
of a negative number by using \texttt{abs(num)} before converting to a
string and then modifying the first digit's sign separately.''
\tcblower
\textbf{Code revision.} An earlier implementation attempts
\texttt{int(str(num)[0])}, which reads the minus sign for negative inputs.
The repair extracts unsigned digits and restores the sign only to the first.
\begin{lstlisting}[style=casepython]
digits = []
abs_num = abs(num)
str_num = str(abs_num)
if num < 0:
    digits.append(-int(str_num[0]))
else:
    digits.append(int(str_num[0]))
for ch in str_num[1:]:
    digits.append(int(ch))
if sum(digits) > 0:
    count += 1
\end{lstlisting}
\textbf{Result.} The final implementation passes the evaluation.
\end{casepanel}

The reflection turns the sign-handling concern into a concrete
operation on the first digit, which the successful code implements.

\Needspace{6\baselineskip}
\subsubsubsection{\textbf{Case 4: Removing All Occurrences of Repeated Values}}
\label{app:case-duplicate-removal}

\textbf{Task.} \texttt{remove\_duplicates} removes every value that
appears more than once and preserves the order of the rest. For example,
$[1,2,3,2,4]$ must become $[1,3,4]$.

\begin{casepanel}{Case 4: Baselines}
\textbf{Reflexion.} A reflection states that ``The code should remove all
elements that occur more than once''. The final program still uses a
\texttt{seen} set to retain the first occurrence of each value.

\textbf{DoT-Bank.} A reflection suggests that ``the test case might be
expecting a different behavior''. The final program still fails to
filter values by frequency.

\textbf{ParamMem.} Reflections recommend including each element only
once, regardless of its frequency. The final program retains repeated
values rather than removing them entirely. All three baselines fail
the evaluation.
\end{casepanel}

\begin{casepanel}[casesuccess]{Case 4: CoMem}
\textbf{Hypotheses.} The guidance warns against ``Incorrectly assuming
that the first occurrence of an element determines its frequency'' and
calls for checking counts before adding elements to the result.

\textbf{Reflection.} ``Track the count of each element using a dictionary
and only include elements in the result if they appear exactly once in
the entire list, not just up to their first occurrence.''
\tcblower
\textbf{Code revision.} The initial program returns $[1,2,3,4]$ for
$[1,2,3,2,4]$. The repair first counts all occurrences in \texttt{count},
then filters the original sequence to return $[1,3,4]$.
\begin{lstlisting}[style=casepython]
for num in numbers:
    if count[num] == 1:
        result.append(num)
return result
\end{lstlisting}

\textbf{Result.} The final implementation passes the evaluation.
\end{casepanel}

The hypotheses warn against relying on the first occurrence. The reflection
turns this concern into a global frequency check, which the revised code
implements.

\subsubsection{LiveCodeBench}
\label{app:case-livecodebench}

We show three tasks in LiveCodeBench where CoMem succeeds after
reflection, with comparisons against Reflexion, DoT-Bank, and ParamMem.
These cases illustrate checks on rotation validity, graph-state updates,
and candidate eligibility.
Each case shows hypotheses alongside the corresponding reflections and
code revisions.


\Needspace{6\baselineskip}
\subsubsubsection{\textbf{Case 5: Validating the Entire Rotation}}
\label{app:case-lcb-rotation}

\textbf{Task.} Given an array of distinct integers, return the minimum
number of right shifts needed to sort it, or $-1$ if no rotation is sorted.
For example, $[3,4,5,1,2]$ requires two right shifts to become
$[1,2,3,4,5]$.

\Needspace{6\baselineskip}
\begin{casepanel}{Case 5: Baselines}
\textbf{Reflexion.} The final code compares the cyclically indexed input
against a sorted copy and computes the required shift count. It passes
all 15 evaluation tests.

\textbf{DoT-Bank.} A reflection recommends that the solution ``thoroughly
validate the sorted condition after identifying the break point''.
The final code still checks only the first decrease and the end-to-start
relation, passing 10 of 15 tests.

\textbf{ParamMem.} The final code constructs the candidate rotation and
compares it with a sorted copy of the array. It passes all 15 tests.
\end{casepanel}

\begin{casepanel}[casesuccess]{Case 5: CoMem}
\textbf{Hypotheses.} The used hypotheses cover whether the array is a
valid rotation of a sorted sequence, the zero-shift case, and the
conversion from a pivot position to the required number of right shifts.

\textbf{Reflection.} ``After identifying the pivot, ensure that the array
is a valid rotation by checking that the last element is less than or
equal to the first element and that all elements after the pivot are
in increasing order.''
\tcblower
\textbf{Code revision.} The earlier code finds the first decrease and
checks the endpoints but can miss further decreases. The repair adds
a scan of the remaining interval.
\begin{lstlisting}[style=casepython]
for i in range(pivot, n - 1):
    if nums[i] > nums[i + 1]:
        return -1
\end{lstlisting}
For illustration, $[3,1,4,2]$ cannot be sorted by rotation. The earlier
code returns 3, while the added check detects the second decrease and
returns $-1$.

\textbf{Result.} The final implementation passes all 15 evaluation tests.
\end{casepanel}

Here, the reflection turns the rotation-validity concern into a check
for additional decreases after the first pivot, which the repair implements.

\subsubsubsection{\textbf{Case 6: Preserving Isolation Counts During Edge Removal}}
\label{app:case-lcb-isolation}

\textbf{Task.} \emph{Isolation} maintains an undirected graph under edge
insertions and removal of all edges incident to a chosen vertex. It must
report the number of isolated vertices after each operation. Deleting
edges from an already isolated vertex should leave this count unchanged.

\begin{casepanel}{Case 6: Baselines}
\textbf{Reflexion.} The final code updates isolation counts as existing
edges are removed and leaves them unchanged when no edge exists. It
passes all 14 evaluation tests.

\textbf{DoT-Bank.} The final code decreases the isolated count when an
endpoint loses its last edge, reversing the required update. It passes
0 of 14 tests.

\textbf{ParamMem.} A reflection recommends: ``Before updating the
isolated\_count, check if the vertex's degree was 0 before the change to
ensure we don't overcount isolated vertices.'' The final code still
increments the count after an empty deletion, passing 1 of 14 tests.
\end{casepanel}

\Needspace{6\baselineskip}
\begin{casepanel}[casesuccess]{Case 6: CoMem}
\textbf{Hypotheses.} The used hypotheses cover accurate degree tracking,
the effect of edge removal on neighboring vertices, and efficient
dynamic graph updates.

\textbf{Reflection.} ``Before removing edges, check if the vertex is
already isolated and adjust the count accordingly.''
\tcblower
\textbf{Code revision.} Deleting edges from one of two isolated vertices
incorrectly returns 3. The repair checks whether the vertex is already
isolated, preserving the correct count of 2.
\par\noindent
\begin{minipage}{\linewidth}
\begin{lstlisting}[style=casepython]
if degree[v] == 0:
    print(isolated_count)
    continue
\end{lstlisting}
\textbf{Result.} The final implementation passes all 14 evaluation tests.
\end{minipage}
\end{casepanel}

Here, the reflection turns the state-tracking concern into a check on
whether the vertex was already isolated, and the repair implements it
before updating the graph.

\Needspace{6\baselineskip}
\subsubsubsection{\textbf{Case 7: Maintaining All Eligible Earlier Elements}}
\label{app:case-lcb-distance}

\textbf{Task.} Minimize $|\mathrm{nums}[i]-\mathrm{nums}[j]|$ subject to
$|i-j|\geq x$. The distance constraint is a lower bound, so sufficiently
old elements remain eligible as the scan advances. Self-comparisons
violate this constraint when $x>0$.

\Needspace{6\baselineskip}
\begin{casepanel}{Case 7: Baselines}
\textbf{Reflexion.} A reflection calls for considering all earlier
elements at least $x$ positions away. A later suggestion returns to
removing old elements, and the final program discards eligible history.

\textbf{DoT-Bank.} Reflections describe the valid range as
``i - x to i''. The final program retains nearby positions and permits
self-comparisons, reversing the distance requirement.

\textbf{ParamMem.} A reflection recommends checking both neighbors of a
binary-search insertion point. The final program nevertheless queries a
collection that already contains the same occurrence being searched for.
\end{casepanel}

\Needspace{6\baselineskip}
\begin{casepanel}[casesuccess]{Case 7: CoMem}
\textbf{Hypotheses.} ``The sliding window approach may miss valid pairs
because it only considers elements within a fixed distance, not all pairs
that are at least x indices apart.''

\textbf{Reflection.} ``The algorithm incorrectly considers elements that
are not at least x indices apart due to improper window management.''
\tcblower
\textbf{Code revision.} At position $i\geq x$, the repair adds the newly
eligible value $\mathrm{nums}[i-x]$ and retains previously eligible
history. It searches for the closest value to $\mathrm{nums}[i]$ in
\[
  \mathcal{H}_i=\{\mathrm{nums}[j]\mid 0\leq j\leq i-x\}.
\]
The maintained candidates now cover all eligible earlier positions,
correcting both the removal of valid history and the inclusion of
positions that are too close.

\textbf{Result.} The final implementation is reported to pass the complete
evaluation, while the compared baselines fail.
\end{casepanel}

The hypothesis questions which valid candidates are excluded, while the
reflection identifies invalid candidates being included. The revision
addresses both concerns by replacing a recent window with accumulated
eligible history.

\subsubsection{MuSiQue}
\label{app:case-musique}

We show two tasks in MuSiQue where CoMem returns the reference answer
after reflection while Reflexion and DoT-Bank return incorrect final
answers. The saved hypotheses cover entity roles, location, and
administrative levels, illustrating several ways to question an initial
interpretation. These excerpts are from saved refined pitfall summaries,
whose correspondence to the exact input at each step is not recorded.


\Needspace{5\baselineskip}
\subsubsubsection{\textbf{Case 8: Distinguishing an Actor from the Character Played}}
\label{app:case-musique-character}

\textbf{Task.} The question asks who the creator of \emph{Push, Nevada}
played in \emph{Dazed and Confused}. The passages identify Ben Affleck as
a co-creator of the show and associate him with Fred O'Bannion in the film.
The required answer is the character's name, \emph{Fred O'Bannion}.

\begin{casepanel}{Case 8: Baselines}
\textbf{Reflexion.} A reflection states:
``The creator of \emph{Push, Nevada} is Ben Affleck, and he played the
role of Fred O'Bannion in \emph{Dazed and Confused}.'' Nevertheless, the
answers alternate between
Ben Affleck and Matthew McConaughey, ending with Ben Affleck.

\textbf{DoT-Bank.} ``The response failed to cross-check
the actor's filmography for accuracy.'' Its answers repeatedly substitute
actor names without returning the required character. The final answer
is ``Matthew McConaughey''.
\end{casepanel}

\begin{casepanel}[casesuccess]{Case 8: CoMem}
\textbf{Hypotheses.} ``Confusing the fictional character with
a real person'' and ``Mixing up the roles of creator and actor'' address
different sources of confusion in the relation between the show, its
creator, and the film role.

\textbf{Reflection.} ``Confusing the creator of
\emph{Push, Nevada} with an actor in \emph{Dazed and Confused}''.
\tcblower
\textbf{Initial answer.} ``Matthew McConaughey''.

\textbf{Revised answer.} ``Fred O'Bannion''.

\textbf{Result.} The revision matches the reference answer exactly and completes the
relation from the show's creator to the character he plays.
\end{casepanel}

Reflexion verbalizes the correct relation but still answers with the
actor's name. DoT-Bank continues to
substitute actors. CoMem's revision returns the character required by
the question, showing the importance of both resolving entities and
preserving the requested answer type.

\subsubsubsection{\textbf{Case 9: Preserving Administrative Levels Across a Reasoning Chain}}
\label{app:case-musique-county}

\textbf{Task.} The question asks for the capital of a county bordering
the county where Pierce Manufacturing is headquartered. The dataset's
annotated chain links the company to Appleton, then to Outagamie County,
neighboring Brown County, and finally \emph{Green Bay}.

\Needspace{6\baselineskip}
\begin{casepanel}{Case 9: Baselines}
\textbf{Reflexion.} ``Appleton is located in Brown County,
Wisconsin.'' This misidentifies the intermediate county in the annotated
chain. Later reflections continue to reason from that error, and the
final answer is ``Shawano''.

\textbf{DoT-Bank.} ``The answer fails to recognize that the
question requires identifying a county, not a city.'' The question asks
for the county's capital, so this changes the required answer type.
The final answer is ``Madison''.
\end{casepanel}

\begin{casepanel}[casesuccess,height fixed for=first]{Case 9: CoMem}
\textbf{Hypotheses.} ``Confusing headquarters location'',
``Incorrectly assuming neighboring counties'', and ``Mixing up
administrative divisions'' identify possible errors in locating the
company and following the county relations.

\textbf{Reflection.} ``The previous answer incorrectly assumed
Pierce Manufacturing's headquarters are in a different county than
Appleton, Wisconsin.''
\tcblower
\begin{minipage}{\linewidth}
\textbf{Initial answer.} ``Madison''.

\textbf{Revised answer.} ``Green Bay''.

\textbf{Result.} The revised answer returns the endpoint of the dataset's annotated
chain and matches the reference answer exactly.
\end{minipage}
\end{casepanel}

Reflexion confuses the intermediate county. DoT-Bank misstates the
requested administrative level. CoMem's reflection takes up the
headquarters-location concern, and its revised answer matches the
annotated endpoint. The supplied passages leave the county-adjacency and
county-seat relations implicit, so the complete chain here follows the
dataset annotation.

\section{AI Use Statement}
\label{app:ai-use-statement}

We used generative AI tools to assist with drafting and polishing the
manuscript based on outlines, technical content, experimental results,
and analyses provided by the authors. This assistance included organizing the supplied
material into prose, refining sentence structure, and improving clarity
and readability. The authors reviewed and revised the AI-assisted text
sentence by sentence and take full responsibility for the final manuscript, including its
claims, results, and references.

\end{document}